\documentclass{article}

\usepackage{arxiv}

\usepackage[utf8]{inputenc} 
\usepackage[T1]{fontenc}    
\usepackage{hyperref}       
\usepackage{url}            
\usepackage{booktabs}       
\usepackage{amsfonts}       
\usepackage{nicefrac}       
\usepackage{cite}
\usepackage{amsthm}
\usepackage[title]{appendix}
\usepackage{IEEEtrantools} 
\usepackage{amsmath,amssymb,amsfonts}
\usepackage{graphicx}
\usepackage{xcolor}
\usepackage{algorithm,algorithmic}
\usepackage{hyperref}
\hypersetup{hidelinks=true}
\usepackage{textcomp}
\usepackage{subcaption}
\usepackage{chngcntr}
\usepackage{apptools}
\newtheorem{theorem}{Theorem}[section]
\newtheorem{assumption}[theorem]{Assumption}
\newtheorem{lemma}[theorem]{Lemma}
\newtheorem{corollary}[theorem]{Corollary}
\newtheorem{proposition}[theorem]{Proposition}
\newtheorem{remark}[theorem]{Remark}
\usepackage{microtype}      
\usepackage{lipsum}
\usepackage{graphicx}
\graphicspath{ {./images/} }

\title{Regularized Q-Learning with Linear Function Approximation}

\author{
 Jiachen Xi \\
  Department of Industrial and Systems
Engineering\\
  Texas A\&M University\\
  College Station, TX 77843 \\
  \texttt{jx3297@tamu.edu} \\
   \And
 Alfredo Garcia \\
  Department of Industrial and Systems
Engineering\\
  Texas A\&M University\\
  College Station, TX 77843 \\
  \texttt{alfredo.garcia@tamu.edu;} \\
  \And
 Petar Mom\v cilovi\'c \\
  Department of Industrial and Systems
Engineering\\
  Texas A\&M University\\
  College Station, TX 77843 \\
  \texttt{petar@tamu.edu} \\}
\date{}
\begin{document}
\maketitle
\begin{abstract}
Regularized Markov Decision Processes serve as models of sequential decision making under uncertainty wherein the decision maker has limited information processing capacity and/or aversion to model ambiguity.
With functional approximation, the convergence properties of learning algorithms for regularized MDPs (e.g. soft Q-learning) are not well understood because the composition of the regularized Bellman operator and a projection onto the span of basis vectors is not a contraction with respect to any norm. In this paper, we consider a bi-level optimization formulation of regularized Q-learning with linear functional approximation. The {\em lower} level optimization problem aims to identify a value function approximation that satisfies Bellman's recursive optimality condition and the {\em upper} level aims to find the projection onto the span of basis vectors.
This formulation motivates a single-loop algorithm with finite time convergence guarantees. The algorithm operates on two time-scales: updates to the projection of state-action values are `slow' in that they are implemented with a step size that is smaller than the one used for `faster' updates of approximate solutions to Bellman's recursive optimality equation. We show that, under certain assumptions, the proposed algorithm converges to a stationary point in the presence of Markovian noise.
In addition, we provide a performance guarantee for the policies derived from the proposed algorithm.
\end{abstract}


\section{Introduction}
Recent literature has examined {\em regularized} Markov Decision Processes MDPs as models of sequential decision making under uncertainty that account for limited information processing capacity \cite{Tishby_2011, Ortega_2013, Matejka_2015} and aversion to model misspecification or ambiguity \cite{Hansen_2018}.
In this literature, low entropy policies are not necessarily optimal because the agent has limited capacity to process new information and/or is averse to model uncertainty. 
A parallel strand of the literature in reinforcement learning algorithms \cite{schulman2015trust,haarnoja2018soft,geist2019theory,yang2019regularized} has also employed equivalent formulations with different motivations, namely that of ensuring learned policies are robust to perturbations in both the dynamics and the reward function \cite{eysenbach2021maximum}.
For example, negative Shannon entropy regularization has been employed in \cite{Haarnoja_2017, haarnoja2018soft} to motivate ``soft'' Q-learning and actor-critic algorithms with robust performance.
Regularization has also been used to ensure learned policies are sparse (e.g., Tsallis entropy \cite{lee2018sparse,martins2016softmax,yang2019regularized}). This approach maintains a multi-modal policy form but restricts the support to only optimal and near-optimal actions, thereby mitigating the impact of detrimental actions. 
Nonetheless, the convergence of regularized Q-learning with linear functional approximation is not guaranteed in general \cite{baird1995residual}. 
This is because the composition of a projection onto the span of predefined features with the {\em regularized} Bellman operator is not necessarily a contraction with respect to {\em any} norm. 
Hence, a regularized Q-learning iteration process may diverge because the objectives of maximizing expected discounted reward and projecting onto the linear basis may ``push and pull'' solution updates in significantly different directions. To our knowledge, there are no algorithms with finite-time guarantees to solve {\em regularized} MDP models with linear function approximation and general strongly convex and bounded regularizer.

To address this gap, we consider a bi-level optimization approach for regularized $Q$-learning that enables us to overcome this challenge. The {\em lower} level optimization problem aims to identify a value function approximation (referred to as the {\em main} solution) that satisfies Bellman's recursive optimality condition. The {\em upper} level aims to find a projection onto the span of basis vectors, i.e. a parametrized linear function (referred to as the {\em target} solution) that best approximates the {\em main} solution in the sense of minimizing mean square error.
This formulation motivates a single-loop algorithm wherein at each iteration both the upper level and the lower level solutions are updated with different stepsizes: updates to the projection of state-action values are `slow' in that they are implemented with a step size that is smaller than the one used for `faster' updates of approximate solutions to Bellman's recursive optimality equation. We demonstrate that in the presence of Markovian noise the proposed algorithm converges to a stationary point at a rate $\mathcal{O}(T^{-1/4}(\log T)^{1/2})$ with $T>0$ denoting the total number of iterations. This result is cast in terms of the $\ell_2$ norm rather than the squared $\ell_2$ norm, because unlike other single-loop
algorithms for reinforcement learning \cite{hong2023two} the proposed algorithm does not require a projection for implementing upper level updates.
Lastly, we establish finite-time performance guarantees for the learned policies. These guarantees reveal that, compared to the optimal policy, the expected performance gap of policies is $\mathcal{O}(T^{-1/4}(\log T)^{1/2})$, augmented by the function approximation error and the error introduced by the smooth truncation operator.

\subsection{Related Work}

The stability of Q-learning and its related variants, particularly in the context of function approximation, has long been a focus of research \cite{meyn2023stability,sutton2018reinforcement}. Numerous techniques have been proposed to address these stability concerns, including the target network \cite{mnih2015human}, double estimator \cite{hasselt2010double}, fitted value iteration \cite{ernst2005tree}, gradient-based approaches \cite{maei2010toward}, Zap Q-learning \cite{chen2020zap,devraj2017zap} and its variant, Zap Zero \cite{meyn2022control} among others. In this subsection, we briefly discuss some works that are most closely related to our work.

The {\em single}-loop algorithm proposed in this paper stands in contrast with {\em nested}-loop algorithms, e.g. \cite{chen2022target, ma2021greedy} which require greater computational effort unless precise fine-tuning of the number of inner-loop iterations is available.
Coupled Q-learning \cite{carvalho2020new} is a single-loop approach that also relies on updates to the main and target solutions with decreasing step-sizes in a two-time scale manner as in \cite{borkar1997stochastic}.
However, as shown in \cite{chen2022target} (see appendix E.3) the performance guarantee for Coupled Q-learning involves significant bias unless the discount factor is sufficiently small.

Our work is also related to approaches to Q-learning with ridge regularization \cite{limregq,zhang2021breaking}. However, ridge regularization results in a modified Bellman
equation with an implicitly scaled down discount factor (see
appendix E.2 in \cite{chen2022target}). Therefore it is not clear how the identified solution approximates the solution of either the original MDP or its regularized version. In contrast, the finite-time guarantee provided in this paper ensures the identified solution approximately minimizes Mean Squared Projected Bellman Error (MSPBE), a measure of the extent to which the identified solution fails to satisfy Bellman's optimality equation.

Finally, the 
Greedy-GQ algorithm \cite{maei2010toward}, inspired by the success of TDC in off-policy evaluation \cite{sutton2009fast}, is arguably the closest to our work.
Greedy-GQ employs a two-timescale gradient-based approach to minimize the MSPBE, and its asymptotic convergence to a stationary point is established over i.i.d. sampled data. Recent work \cite{wang2020finite} provides a finite-time guarantee for the convergence of Greedy-GQ under Markovian noise with a rate of $\mathcal{O}(T^{-1/3} \log T)$, and it has been improved to $\mathcal{O}(T^{-1/2} \log T)$ in \cite{wang2022finite}.
Subsequent work \cite{xu2021sample} further improves the result to $\mathcal{O}(T^{-1/2})$ with the use of a mini-batch gradient.

The performance guarantees studied in these Greedy-GQ papers concern the solution to a {\em standard} MDP problem (i.e. not regularized). In contrast, in this paper, we consider the solution of {\em regularized} MDP problem wherein the regularizer belongs to a general class of strongly convex and bounded functions. This class includes negative Shannon and negative Tsallis entropy functions. 
Such regularized MDPs are used to model sequential decision making under uncertanity with limited information processing capacity \cite{Tishby_2011, Ortega_2013, Matejka_2015} and/or aversion to model misspecification or ambiguity \cite{Hansen_2018}.
To our knowledge, there are no algorithms with finite-time guarantees to solve such {\em regularized} MDP models with linear function approximation. Our work contributes to fill this gap.

\section{Preliminaries}
In this section, we present the preliminaries for regularized Q-learning. We denote the dot product as $\langle\cdot,\cdot\rangle$. The $\ell_2$ and $\ell_\infty$ norms are denoted by $\|\cdot\|$ and $\|\cdot\|_{\infty}$, respectively.
\subsection{Markov Decision Processes (MDPs)}
We consider an infinite-horizon regularized Markov Decision Processes (MDPs) with finite action space. An MDP can be described as a tuple $\mathcal{M} \triangleq \left(\mathcal{S},\mathcal{A},P,\mu_0,R,\gamma\right)$, where $\mathcal{S}$ denotes the state space and $\mathcal{A}$ denotes the action space. $P$ is the dynamics: $P(s^\prime\mid s,a)$ is the transition probability from state $s$ to $s^\prime$ by taking action $a$. $\mu_0$ is the initial distribution of the state. The reward function, $R:\mathcal{S}\times\mathcal{A} \rightarrow \mathbb{R}$, is bounded in absolute value by $R_{\max}\geq 0$, i.e., $-R_{\max} \leq R(s,a)\leq R_{\max}$ for all $(s,a)\in\mathcal{S}\times\mathcal{A}$. $\gamma \in (0,1)$ is the discount factor. A policy $\pi: \mathcal{S} \rightarrow \Delta(\mathcal{A})$ is defined as a mapping from the state space $\mathcal{S}$ to the probability distribution over the action space $\mathcal{A}$. For discrete time $t\geq 0$, the trajectory, starting from an initial state $s_0\sim \mu_0(\cdot)$, generated by the policy $\pi$ in the MDP $\mathcal{M}$ can be represented as a set of transition tuples $\left\{s_t,a_t,s_{t}^\prime\right\}_{t\geq 0}$, where $a_t\sim \pi(\cdot\mid s_t)$, $s_t^\prime\sim P(\cdot \mid s_t,a_t)$, and $s_{t+1} = s_{t}^\prime,\forall t\geq 0$. 

The state value and state-action value functions corresponding to a policy $ \pi $ are defined by the expected cumulative rewards obtained when initiating from state $ s \in \mathcal{S}$ or state-action pair $ (s, a) \in \mathcal{S}\times \mathcal{A}$, and subsequently adhering to policy $ \pi $, respectively:
\[
V_{\pi}(s)\triangleq \mathbb{E}\big[\sum_{t=0}^\infty \gamma^{t}R\left(s_t,a_t\right)\bigg\vert s_0 = s,P,\pi\big],
\]
and $ 
Q_{\pi}(s,a)= R(s,a)+\gamma\mathbb{E}_{s'\sim P(\cdot|s,a)}[V_{\pi}(s')]$.
In general, the objective of an MDP is to identify the optimal policy $ \pi^* $ that maximizes the expected cumulative discounted reward. Mathematically, this can be expressed as:
\[
\pi^* \in \arg\max_{\pi} \mathbb{E}_{s \sim \mu_0}\left[V_{\pi}(s)\right].
\]
 Let $Q: \mathcal{S}\times\mathcal{A}\rightarrow \mathbb{R}$. The Bellman operator is defined as: 
\begin{align}
\mathcal{B}Q(s,a)\triangleq R(s,a)+\gamma \mathbb{E}_{s'\sim P(s,a)}\big[\max_{a\in \mathcal{A}}\{Q(s',a)\}\big].
\label{Bellman}
\end{align}
Under certain assumptions, this operator is a contraction with modulus $\gamma<1$ and its unique fixed point $Q^*$ characterizes the optimal value function as follows:
\[
V_{\pi^*}(s) =\max_{a\in\mathcal{A}} Q^*(s,a), \quad \forall s\in \mathcal{S}.
\]
\subsection{Regularized Markov Decision Process}
Let $G: \Delta(\mathcal{A}) \rightarrow \mathbb{R}$ be a strongly convex function, that is bounded, i.e. there exists $B >0$ such that $-B \leq G(p) \leq B$ for all $p\in \Delta(\mathcal{A})$. 
For a given policy $\pi$ , the {\em regularized} value function is defined as
\begin{align*}
    V_{\pi,G_\tau}(s)  & \triangleq \mathbb{E}\big[\sum_{t=0}^\infty \gamma^{t}\big(R\left(s_t,a_t\right) - G_{\tau}(\pi(\cdot\mid s_t))\big)\big \vert s_0 = s,P,\pi\big]
\end{align*}
where $G_\tau \triangleq \tau \cdot G$, and $\tau > 0$ is a positive coefficient that determines the degree of regularization applied. Let us also define $ 
Q_{\pi,G_{\tau}}(s,a)\triangleq R(s,a)+\gamma \mathbb{E}_{s'\sim P(\cdot|s,a)}[V_{\pi,G_{\tau}}(s')]$.

The {\em regularized} Bellman operator is defined as follows:
\begin{equation*}
\mathcal{B}_{G_{\tau}}Q(s,a)\triangleq R(s,a)\\+\gamma \mathbb{E}_{s'\sim P(s,a)}\big[\max_{\pi}\big\{\mathbb{E}_{a' \sim \pi(\cdot|s')} [Q(s',a')]-G_{\tau}(\pi(\cdot|s'))\big\}\big].
\label{Soft_Bellman}
\end{equation*}
Note that without regularization (i.e. $\tau=0$), definitions \eqref{Bellman} and \eqref{Soft_Bellman} are equivalent. The regularized Bellman operator can be re-written as:
\begin{align}
\mathcal{B}_{G_{\tau}}Q(s,a)= R(s,a)+\gamma \mathbb{E}_{s'\sim P(s,a)}\big[
G^*_{\tau}(Q(s',\cdot))\big]
\end{align}
wherein $G^*_{\tau}$ is the convex conjugate of $G_{\tau}$, i.e.:
\[G^*_{\tau}(Q(s',\cdot))\triangleq
\max_{\pi}\big\{\mathbb{E}_{a' \sim \pi(\cdot|s')} [Q(s',a')]-G_{\tau}(\pi(\cdot|s'))\big\}
\]

The {\em regularized} Bellman operator is a contraction in $\ell_{\infty}$ norm with modulus $\gamma<1$ (see Proposition 2 in \cite{geist2019theory}). Let $V^*_{G_{\tau}}(s) \triangleq \max_{\pi} \mathbb{E}_{s \sim \mu_0} \big[V_{\pi,G_\tau}(s)\big]$.
According to Theorem 1 in \cite{geist2019theory}, the unique fixed point of the regularized Bellman operator, say $Q^*_{G_\tau},$ characterizes the optimal value in the following sense: 
\[
G^*_{\tau}(Q^*_{G_{\tau}}(s,\cdot))=\max_{\pi} V_{\pi,G_\tau}(s)
\]
{\bf Some Examples of Regularizers}:
A common choice for regularizer is negative Shannon entropy. As mentioned before, this choice has been used as a way to account for information processing costs \cite{Tishby_2011, Ortega_2013, Matejka_2015} and to motivate the ``soft'' Q-learning algorithm. 
In this case $G_{\tau}(\pi(\cdot|s')) = \tau\sum_{a \in A} \pi(a|s') \log \pi(a|s')$ and
\begin{align*}
 G^*_{\tau}(Q(s',\cdot)) = \tau \log \big(\sum_{a\in A} \exp{\frac{1}{\tau}Q(s',a)}\big)  
\end{align*}
Another choice of regularizer is the negative Tsallis entropy, $G(\pi(\cdot|s')) = \frac{1}{2}(\left\|\pi(\cdot|s')\right\|^2-1)$. In \cite{lee2018sparse}, the authors show that this regularizer induces a sparse and multi-modal optimal policy.
Some additional properties of regularized MDPs are stated in the following proposition (see Proposition 1 in \cite{geist2019theory} for proofs).
\begin{proposition}\label{prop: regularizer}
Let $G$ be a strongly convex function bounded by $B >0$ and $\tau >0$ be the regularization coefficient associated with $G$. The following hold:
\begin{enumerate}
\item[i.] The conjugate function is bounded: 
\begin{align*}
\Big|G^*_\tau\big(Q(s,\cdot)\big) - \max_{a \in \mathcal{A}}Q(s,a)\Big| &\leq \tau B
\end{align*}
\item[ii.] The gradient of the convex conjugate can be caharacterized as follows:
\begin{align*}
\nabla G_\tau^*(Q(s,\cdot)) &=\arg\max_{\pi}\big\{\mathbb{E}_{a \sim \pi(\cdot|s)} [Q(s,a)]-G_{\tau}(\pi(\cdot|s))\big\}
\end{align*}
\item[iii.] (Lipschitz gradients) For some constant $L_G>0$.

{\small
\begin{align*}
\left\|\nabla G_\tau^*(Q_1(s,\cdot)) - \nabla G_\tau^*(Q_1(s,\cdot))\right\| &\leq \frac{L_G}{\tau}\left\|Q_1(s,\cdot)-Q_2(s,\cdot)\right\|
\end{align*}
}
\item[iv.] The optimal policy $\pi^*_{\tau}(\cdot|s) = \nabla G^*_{\tau}\left(Q^*_{\tau}\left(s,\cdot\right)\right)$, for all $s\in\mathcal{S}$.
\end{enumerate}
\end{proposition}

\subsection{Linear Function Approximation}
Function approximation techniques are prevalent as they offer an effective solution to this problem by allowing agents to approximate the values using significantly fewer parameters. In this study, we focus on leveraging linear function approximation which approximates the value functions linearly in the given basis vectors. 

The basis vectors $\phi_i\in \mathbb{R}^{|\mathcal{S}||\mathcal{A}|}$, $i = 1,2,\dots,d$ are, without loss of generality, linearly independent. We denote $\phi(s,a) = \left[\phi_1(s,a),\phi_2(s,a),\dots,\phi_d(s,a)\right]^\top\in \mathbb{R}^d$ for all $(s,a)$ and let $\Phi \in \mathbb{R}^{|\mathcal{S}||\mathcal{A}|\times d}$ defined by 
$\Phi = \left[\phi(s_1,a_1), \dots, \phi(s_{|\mathcal{S}|},a_{|\mathcal{A}|})
   \right]^\top$.
We approximate state-action function as $\hat{Q}_{\theta}\triangleq \Phi\theta$, where $\theta \in \mathbb{R}^d$ is the parameter vector. We also use $\hat{Q}_{\theta}(s,\cdot) = \phi(s,\cdot)^\top\theta$ to denote the vector of the approximated state-action values of all actions given the state $s\in \mathcal{S}$, where $\phi(s,\cdot) = \left[\phi(s,a_1),\dots,\phi(s,a_{|\mathcal{A}|})\right]$.
By Proposition 2.2 (iv), the optimal policy satisfies: 
\begin{align}
\pi_{\theta}(\cdot|s)&= \nabla G^*_{\tau}\big(\hat{Q}_{\theta}(s,\cdot)\big)
\label{eq: pi_def}
\end{align}

Suppose $\mu$ is a distribution that exhibits full support across the state-action space, let $D_{\mu}\in \mathbb{R}^{|\mathcal{S}||\mathcal{A}|\times |\mathcal{S}||\mathcal{A}|}$ be a diagonal matrix with the distribution $\mu$ on its diagonal. Furthermore, we define the weighted $\ell_2$-norm with respect to the distribution $\mu$ as $\|x\|_{D_{\mu}}\triangleq \sqrt{x^\top D_{\mu}x}$. Let $\Pi_{D_{\mu}}$ be the projection operator which maps a vector onto the span of the basis functions with respect to the norm $\|\cdot\|_{D_{\mu}}$; it is given by $\Pi_{D_{\mu}} = \Phi\left(\Phi^\top D_{\mu}\Phi\right)^{-1} \Phi^\top D_{\mu}$ under the assumption that $\Phi^\top D_{\mu}\Phi$ is a positive definite matrix. 

Within the framework of linear function approximation, the primary objective of value function estimation is to determine an optimal parameter $\theta^*$ that satisfies the regularized projected Bellman equation:
$\hat{Q}_{\theta^*} = \Pi_{D_{\mu}}\mathcal{B}_{\tau}\hat{Q}_{\theta^*}$,
where $\mu$ is given.
While the projection operator $\Pi_{D_{\mu}}$ is known to be a weighted $\ell_2$ non-expansive mapping, it is important to note that the composite operator $\Pi_{D_{\mu}}\mathcal{B}_{\tau}$ does not generally exhibit contraction properties with respect to any norm \cite{chen2022target}.

\subsection{Standing Assumptions}
We now state the standing assumptions for our analysis. Let $\pi_{\mathrm{bhv}}$ be the behavioral policy used to collect data in regularized MDP $\mathcal{M}_{\tau}$ and we adopt the following assumption. 
\begin{assumption}\label{assump: Ergodicity}
    {\em The behavioral policy $\pi_{\mathrm{bhv}}$ satisfies $\pi_{\mathrm{bhv}}(a\mid s) >0$ for all $(s, a)$, and the Markov chain induced by it is irreducible and aperiodic. Then there exist constants $\kappa>0$ and $\rho \in (0,1)$ such that
    \[
    \sup _{s_0 \in \mathcal{S}} d_{T V}\left(\mathbb{P}\left(\left(s_t, a_t\right) \in \cdot \mid s_0, \pi_{\mathrm{bhv}},P\right), \mu_{\mathrm{bhv}} \right) \leq \kappa \rho^t, \quad \forall t,
    \]
    where $d_{TV}$ is the total-variation distance and $\mu_{\mathrm{bhv}}$ is the stationary distribution of state-action pairs. }
\end{assumption}
Assumption \ref{assump: Ergodicity} guarantees the ergodicity of the Markov chain, which in turn ensures that it converges to its stationary distribution at a geometric rate. 
Furthermore, Let $\mathcal{D}$ be the stationary distribution of the transition tuples induced by $\pi_{\mathrm{bhv}}$. Consequently, $\mathcal{D} = \mu_{\mathrm{bhv}}\otimes P$, where $\otimes$ denotes the tensor product between two distributions. We use the term $\mathcal{D}$ to also refer to the marginal distribution of state-action pairs, i.e., $\mathcal{D}(s,a) = \mu_{\mathrm{bhv}}(s,a)$ for all $(s,a) \in \mathcal{S}\times\mathcal{A}$, with a slight abuse of notation. Finally we make the following standing assumption 
\begin{assumption}\label{assump: finite_feature}
    $\left\|\phi(s,a)\right\|_2  \leq 1, \forall (s,a)\in \mathcal{S}\times \mathcal{A}$.
\end{assumption}

Note assumption 2.3 can be ensured by normalizing the features (see~\cite{bhandari2018finite,shen2020asynchronous,wang2020finite}). With a slight abuse of notation in the remainder of the paper we use $\Pi$ to represent $\Pi_{D_{\mu_{\mathrm{bhv}}}}$ and $\|\cdot\|$ to refer to $\|\cdot\|_{D_{\mu_{\mathrm{bhv}}}}$.
\section{Problem Formulation}
\subsection{Smooth Truncation Operator}
Motivated by the stabilization of using the truncation operator within linear function approximation as highlighted in \cite{chen2022target}, we explore a smooth variant of this operator. This adaptation maintains the smoothness of Bellman backups in regularized MDPs and is particularly advantageous for gradient-based algorithms, which are our primary focus.
For a threshold $ \delta >0$, we define the smooth truncation operator $ \mathcal{K} $ as
\[
\mathcal{K}_{\delta}(x) \triangleq \delta \cdot \tanh\left(\frac{x}{\delta}\right), \quad \forall x \in \mathbb{R},
\]
where $ \tanh $ is the hyperbolic tangent function; the particular choice of the operator is not essential -- alternative choices are feasible. 
Consequently, $ \mathcal{K}_{\delta} $ maps any value in $\mathbb{R} $ to the interval $ (-\delta, \delta) $. We define $\hat{G}_{\tau,\delta}$ as the composite operator formed by applying the smooth truncation operator $\mathcal{K}_{\delta}$ to the convex conjugate $G_{\tau}^*$ of the regularizer, i.e. 
\[
\hat{G}_{\tau,\delta} (q) = \mathcal{K}_{\delta}\left( G_{\tau}^*\left(q\right)\right), \forall q\in \mathbb{R}^{|\mathcal{A}|}.
\]
In addition, we let $\mathcal{B}_{\tau,\delta}$ be the smooth truncated optimal regularized Bellman operator which is given as, for an arbitrary mapping $Q: \mathcal{S}\times \mathcal{A} \rightarrow \mathbb{R}$:
\[
\mathcal{B}_{\tau,\delta}Q(s,a) \triangleq R(s,a) + \gamma \mathbb{E}_{s^\prime\sim P(\cdot \mid s,a)}\left[\hat{G}_{\tau,\delta} \left(Q\left(s^\prime,\cdot\right)\right)\right],
\]
which preserves the property of $\gamma$-contraction in $\ell_{\infty}$-norm.

In contrast to the hard truncation operator $ \lceil x \rceil_{\delta} \triangleq \max\{\min\{x, \delta\}, -\delta \}$, $ \mathcal{K}_{\delta}$ is differentiable everywhere on $\mathbb{R}$, with the gradient given by
\[
\nabla \mathcal{K}_{\delta}(x) = 1 - \tanh^2\left(\frac{x}{\delta}\right) = 1 - \frac{\mathcal{K}^2_{\delta}(x)}{\delta^2},
\]
which equals $1$ at $x = 0$ and approaches $0$ as $|x| \to \infty$. It is evident that $|\mathcal{K}_{\delta}(x)| \leq |\lceil x \rceil_{\delta}| \leq |x|$ for all $x\in\mathbb{R}$. This property can be beneficial in mitigating the overestimation issue. However, there is a trade-off involving the gap between $\mathcal{K}_{\delta}(x)$ and $ x $ when $x$ is away from the origin. On the other hand, as the value of $ |\frac{x}{\delta}| $ decreases, $ \mathcal{K}_{\delta}(x) $ approaches closer to $ x $. Therefore, we have the inequality
\begin{equation}\label{eq: truncation_gap_inequality}
    \left\vert x-\mathcal{K}_{\delta}(x)\right\vert \leq \left\vert y-\mathcal{K}_{\delta}(y)\right\vert \text{ if } |x|\leq |y| \in \mathbb{R}.
\end{equation}
Further investigation of the threshold $\delta$ is presented in Section~\ref{sec: threshold}.

\subsection{ A Bi-level Formulation for Bellman Error Minimization}
In what follows, we aim to minimize the Mean Squared Projected Bellman Error (MSPBE) defined as
\[
J(\theta) \triangleq \frac{1}{2}\mathbb{E}_{\mathcal{D}} \left(\Pi\mathcal{B}_{\tau,\delta}\hat{Q}_{\theta}(s,a)-\hat{Q}_{\theta}(s,a)\right)^2.
\]

A bi-level optimization formulation for minimizing MSPBE is as follows:
\begin{equation}\label{eq: bi}
\begin{aligned}
\quad \min_{\theta\in \mathbb{R}^d}  f\left(\theta,\omega^*(\theta)\right)
&\triangleq \frac{1}{2}\mathbb{E}_{\mathcal{D}} \left(\phi(s,a)^\top \omega^*\left(\theta\right) - \phi(s,a)^\top\theta\right)^2\\
   \textit{s.t.}\quad \omega^*\left(\theta\right) &\in \arg\min_{\omega\in \mathbb{R}^d} g(\theta,\omega),
\end{aligned}
\end{equation}
where the lower-level objective is 
\[
 g(\theta,\omega) \triangleq \frac{1}{2}\mathbb{E}_{\mathcal{D}} \left(\phi(s,a)^\top \omega - R(s,a) -\gamma \hat{G}_{\tau,\delta}(\hat{Q}_{\theta}\left(s^\prime,\cdot\right)) \right)^2. 
\]
Note that $\hat{Q}_{\omega^*(\theta)}= \Phi\omega^*(\theta)$ represents the projection of $\mathcal{B}_{\tau,\delta}\hat{Q}_{\theta}$ onto the space spanned by the basis functions with respect to the norm $\|\cdot\|_{D_{\mu_{\mathrm{bhv}}}}$, i.e. $\hat{Q}_{\omega^*(\theta)} = \Pi\mathcal{B}_{\tau,\delta}\hat{Q}_{\theta}$. 
Hence, it can be readily seen that the upper level objective function is MSPBE, i.e.
$J\left(\theta\right)= f\left(\theta,\omega^*(\theta)\right)$.
In the above formulation, the parameters $\omega$ are referred to as the {\em main} solution (or network) whereas $\theta$ represents the parameters of the {\em target} solution (or network).

We now show the lower level objective function $g(\theta,\omega)$ is strongly convex in $\omega$:
\begin{proposition}\label{prop: g_convex}
  Under Assumption \ref{assump: Ergodicity}, there exists a constant $\lambda_g >0$ that lower bounds the eigenvalue of $\Sigma\triangleq\mathbb{E}_{\mathcal{D}}[\phi(s,a)\phi(s,a)^\top]$. Consequently,
  $g(\theta,\omega)$ is strongly convex in $\omega$ for any $\theta \in \mathbb{R}^d$, and with modulus $\lambda_g/2$. That is, for any $\omega_1,\omega_2\in \mathbb{R}^d$, $\theta\in\mathbb{R}^d$,
\begin{align*}
g(\theta,\omega_1) &\geq g(\theta,\omega_2) + \left\langle\nabla_{\omega}g(\theta,\omega_2) , \omega_1-\omega_2\right\rangle + \frac{\lambda_g}{2}\|\omega_1-\omega_2\|^2.
\end{align*} 
\end{proposition}
See the proof in Appendix \ref{proof: prop_g_convex}.
In addition, the solution $\omega^*(\theta)$ of the lower level problem has the following closed-form:
\begin{equation}\label{eq:omega_closed_form}
\!\!\!    \omega^*(\theta) = \Sigma^{-1}\mathbb{E}_{\mathcal{D}}\!\left[\phi(s,a)\left(\! R(s,a) + \gamma \hat{G}_{\tau,\delta}(\hat{Q}_{\theta}\left(s^\prime,\cdot\right))\right)\right].
\end{equation}

\subsection{Gradient of Bellman Error}
The gradient of $J(\theta)$ can be derived by using the chain rule as
\begin{align}\label{eq: J_grad}
    \nabla J(\theta) &= \nabla_\theta f(\theta,\omega^*(\theta))-\nabla^2_{\theta\omega}g(\theta,\omega^*(\theta))\left[\nabla^2_{\omega\omega}g(\theta,\omega^*(\theta))\right]^{-1}\nabla_{\omega}f(\theta,\omega^*(\theta))\nonumber\\
    &= \hat{\Sigma}_{\tau,\delta,\theta}\left(\omega^*(\theta) - \theta\right),
\end{align}
where
\begin{align}\label{eq: Sigma_hat}
\hat{\Sigma}_{\tau,\delta,\theta}\triangleq \nonumber\mathbb{E}_{\mathcal{D}}\Big[\Big(\gamma z\left(s^\prime ; \tau,\delta,\theta\right)\sum_{a^\prime \in \mathcal{A}}\pi_{\theta}\left(a^\prime\mid s^\prime\right)\phi\left(s^\prime,a^\prime\right)
-\phi(s,a)\Big)\phi(s,a)^\top\Big],
\end{align} 
and
\begin{equation}
      z\left(s ; \tau,\delta,\theta\right) \triangleq 1- \frac{1}{\delta^2}\hat{G}^2_{\tau,\delta}(\hat{Q}_{\theta}(s,\cdot)),\quad\forall s\in \mathcal{S}.
\end{equation}
Observe that the range of $z$ is $(0,1]$. 
We make the following regularity assumption on the gradient:
\begin{assumption}\label{assump: non-singular}
{\em For $\theta\in\mathbb{R}^d$, the matrix $\hat{\Sigma}_{\tau,\delta,\theta}$ is non-singular. Therefore, there exists a constant $\sigma_{\min}>0$ that lower bounds the smallest singular value of it for all $\theta \in \mathbb{R}^d$.}
\end{assumption}

Note that Assumption \ref{assump: non-singular} is as strong as similar assumptions in \cite{maei2010toward,chen2019performance,devraj2017zap,lee2019unified,melo2008analysis,xu2019two,xu2021sample}. This is because $z\left(s; \tau,\delta,\theta\right) \rightarrow 1$ as $\delta \rightarrow +\infty$. Nonetheless, when $\delta$ has a finite value, Assumption \ref{assump: non-singular} is weaker, especially for a small value of $\delta$.

\section{A Single-Loop Algorithm}
In this section, we introduce a {\em single-loop} algorithm for solving the bi-level problem \eqref{eq: bi} wherein at every iteration the upper level and the lower solutions are updated based upon samples from a {\em surrogate} gradient of the upper level objective and the gradient of the lower level objective.
Specifically, the gradient of the lower level objective is:
\begin{align*}
   \nabla_{\omega} g(\theta,\omega)&=
   \mathbb{E}_{\mathcal{D}}\big[\phi(s,a)\big(\phi(s,a)^\top \omega - R(s,a) 
   -\gamma  \hat{G}_{\tau,\delta}(\hat{Q}_{\theta}\left(s^\prime,\cdot\right))\big)\big] 
\end{align*}
A {\em surrogate} gradient for the upper level objective:
\begin{equation}
\begin{aligned}
    \bar{\nabla}_{\theta}f(\theta,\omega) \triangleq \hat{\Sigma}_{\tau,\delta,\theta}\left(\omega - \theta\right).
\end{aligned}
\end{equation}
wherein $\omega^*(\theta)$ is replaced by $\omega$ in \eqref{eq: J_grad}.
Samples for the upper-level surrogate gradient and lower-level gradient can be readily obtained as follows.

\begin{algorithm}[t!]
  \caption{Single-loop Regularized Q-learning}
  \label{algo}
  \begin{algorithmic}[1]
     \STATE \textbf{Input:} Constant $T$, step sizes $\alpha,\beta$, initial parameters $\omega^0,\theta^0$, behavioral policy $\pi_{\mathrm{bhv}}$.
    \STATE Sample Initial State: $s_0 \sim \mu_0(\cdot)$.
    \FOR {$t = 0,\dots,T-1$}
    \STATE Sample Action and Subsequent State: 
    \item[] \quad $a_t\sim \pi_{\mathrm{bhv}}(\cdot\mid s_t),s_t^\prime \sim P(\cdot \mid s_t,a_t)$.
    \STATE Estimate Lower-Level Gradient: 
    \item[] \quad compute $h_g^t$ by performing \eqref{eq:g_grad}.
    \STATE Update Lower-Level Parameter: perform \eqref{eq: updating_omega}.
    \STATE Estimate Upper-Level Gradient: 
    \item[] \quad compute $h_f^t$ by performing \eqref{eq:f_grad}.
    \STATE Update Upper-Level Parameter: perform \eqref{eq: updating_theta}.
    \STATE Update State: $s_{t+1} = s_t^\prime$.
    \ENDFOR
  \end{algorithmic}
\end{algorithm}

Let $(s_t,a_t,s_t^\prime)$ denote the $t$-th transition on the Markov chain induced by $\pi_{\mathrm{bhv}}$ on MDP $\mathcal{M}_{\tau}$. The action $a_t$ is sampled from $\pi_{\mathrm{bhv}}(\cdot\mid s_t)$ and the next state $s_{t+1} = s_t^\prime$, which is drawn from $P(\cdot\mid s_t,a_t)$. The stochastic gradient of the lower-level problem and the stochastic surrogate gradient of the upper-level problem are represented by:
\begin{equation}\label{eq:g_grad}
    h_g^t \triangleq \phi(s_t,a_t)\left(\phi(s_t,a_t)^\top\omega^t - R(s_t,a_t) -\gamma  \hat{G}_{\tau,\delta}(\hat{Q}_{\theta^t}\left(s^\prime_t,\cdot\right))\right)
\end{equation}
and
\begin{align}
    h_f^t \triangleq \Big(\gamma z_t\sum_{a^\prime\in\mathcal{A}} &\pi_{\theta^t}\left(a^\prime\mid s^\prime_t\right)\phi(s_t^\prime,a^\prime)
    -\phi(s_t,a_t)\Big) \nonumber \\
&\times \phi(s_t,a_t)^\top\left(\omega^{t+1}-\theta^t\right), \label{eq:f_grad}
\end{align}
respectively, where $z_t:=z\big(s_t^\prime; \tau,\delta,\theta^t\big)$. Recall the definition of $\pi_{\theta}$ in \eqref{eq: pi_def}, which represents the optimal policy derived from the state-action value estimation of the target network, $\hat{Q}_{\theta}$.
The parameters are updated with the step sizes $\alpha$ and $\beta$ in the following sense:
\begin{subequations}
\begin{align}
        \omega^{t+1} &= \mathcal{P}\left(\omega^t - \beta h_g^t\right),&\label{eq: updating_omega}\\
\theta^{t+1} &= \begin{cases}
    \theta^t - \frac{\alpha}{\|\theta^t-\omega^{t+1}\|} h_f^t \quad &\text{if } \theta^{t} \ne \omega^{t+1}, \\
    \theta^t\quad &\text{otherwise,}
\end{cases}\label{eq: updating_theta}
\end{align}
\end{subequations}
where $\mathcal{P}$ denotes the projection onto $\ell_2$-balls with a radius of $(R_{\max} + \gamma \delta)/\lambda_g$. This value ensures that the optimal point $\omega^*(\theta^t)$ of the lower-level problem lies within the specified ball. The algorithm for regularized Q-learning with adaptive stepsize is summarized in Algorithm \ref{algo}.

{\bf Remark:}
Utilizing the projection $\mathcal{P}$ constrains the main network, preventing it from taking overly large steps in the incorrect direction \cite{shen2020asynchronous,xu2020non,xu2019two}.
The update rule for the upper-level problem in Algorithm~\ref{algo} \textit{normalizes} the stochastic gradient by dividing it by the difference between the weights of the main and target networks. 

{\begin{lemma}\label{lemma: stability}(Asymptotic stability.) Suppose Assumptions \ref{assump: Ergodicity}, \ref{assump: finite_feature} and \ref{assump: non-singular} hold. The approximated Q-value remains bounded almost surely for all $a\in \mathcal{A}$:
\[
\sup_t\max_{a\in\mathcal{A}} \left\vert \hat{Q}_\theta^t(s_{t+1},a)\right\vert < \infty\text{, w.p.1.}
\]
\end{lemma}
See the proof in Appendix \ref{proof: lemma_stability}.}

\section{Finite-Time Guarantees}\label{section: analysis}
In this section, we present the finite-time convergence guarantee for Algorithm \ref{algo} and a finite-time performance bound of the derived policies.
The approach we take is similar to that of \cite{Wu_2020,hong2023two} wherein the two time scales refer to different constant step sizes for lower level and upper level updates, i.e. $\alpha < \beta$ and the finite-time guarantee is expressed in terms of a target number of iterations $T\geq1$. This differs from other approaches in which the step-sizes {\em converge} to zero at different rates as in \cite{borkar1997stochastic,borkar2018concentration, Bhatnagar_2015}.

\subsection{Convergence Analysis}
The upper level objective function in \eqref{eq: bi} is not smooth. We consider the following relaxation of smoothness:
\begin{lemma}\label{lemma: J_smoothness}
    (Relaxation of smoothness.) Suppose Assumptions \ref{assump: Ergodicity}, \ref{assump: finite_feature} hold, for all $\theta_1,\theta_2 \in \mathbb{R}^d$, then
    \begin{equation*}
        \left\|\nabla J\left(\theta_1\right)-\nabla J\left(\theta_2\right)\right\|\leq \left(L_0 + L_1\left\|\omega^*\left(\theta_2\right)-\theta_2\right\|\right)\left\| \theta_1 -\theta_2\right\|,
    \end{equation*}
    where $L_0 = 4/\lambda_g$ and $L_1 = L_{G}|\mathcal{A}|/\tau + 2/\delta$. Moreover, the following holds:
    \begin{align*}
        J\left(\theta_1\right)- J\left(\theta_2\right) &- \left\langle \nabla J\left(\theta_2\right), \theta_1-\theta_2\right\rangle\nonumber\\
        & \ \ \ \leq \frac{L_0 + L_1\left\|\omega^*\left(\theta_2\right)-\theta_2\right\|}{2}\left\| \theta_1 -\theta_2\right\|^2.
    \end{align*}
\end{lemma}
We refer the readers to Appendix \ref{proof: lemma_J_smoothness} for detailed proof.
A similar property termed $(L_0,L_1)$-smoothness is incorporated into the optimization algorithm, enhancing the speed of resolution, as demonstrated in \cite{zhang2019gradient,zhang2020improved}. By employing the smooth truncation operator to impose bounds and integrating the projection operation as illustrated in equation \eqref{eq: updating_omega}, we ensure that the $\ell_2$-norm of the stochastic gradient of the lower-level problem is constrained.
\begin{lemma}\label{lemma: bound_hg}
    Under Assumption \ref{assump: finite_feature}, we have
    \[
    \left\|h_g^t\right\| \leq \frac{2\left(R_{\max}+\delta\right)}{\lambda_g},\ \text{for } t= 0,1,\dots,T-1.
    \]
\end{lemma}
The detailed proof can be found in Appendix \ref{proof: lemma_bound_hg}.

We present the finite-time convergence guarantees of Algorithm \ref{algo} in the following theorem.

\begin{theorem}\label{theorem: convergence}
Suppose Assumptions \ref{assump: Ergodicity}, \ref{assump: finite_feature} and \ref{assump: non-singular} hold. Let $\alpha = \alpha_0T^{-3/4}, \beta=\beta_0 T^{-1/2}$, where $\alpha_0,\beta_0$ are positive constants, and

\begin{align*}
    \alpha \leq \frac{\beta}{\frac{\log(\beta/\kappa)}{\log \rho}+2},
    \beta \leq&  \frac{\sigma_{\min}}{4},\\
    64\beta+16\beta^2 +8\alpha^2 \leq&   \frac{\sigma_{\min}^2}{L_1}.
\end{align*}
Then, Algorithm \ref{algo} yields
\begin{subequations}
\begin{equation}\label{eq: theorem_bound_1}
        \frac{1}{T}\sum_{t=0}^{T-1}\mathbb{E} \left\|\omega^*(\theta^{t})-\omega^{t+1}\right\| = \mathcal{O}\left(T^{-1/4}(\log T)^{1/2}\right)
\end{equation}
and 
\begin{equation}\label{eq: theorem_bound_2}
\frac{1}{T}\sum_{t=0}^{T-1}\mathbb{E} \left\|\bar{\nabla} f\left(\theta^t,\omega^{t+1}\right)\right\| = \mathcal{O}\left(T^{-1/4}(\log T)^{1/2}\right).
\end{equation}
\end{subequations}
Consequently, the average $\ell_2$-norm of the gradient of the objective function satisfies 
\begin{equation}\label{eq: theorem_bound_3}
\frac{1}{T}\sum_{t=0}^{T-1}\mathbb{E} \left\|\nabla J\left(\theta^t\right)\right\| =\mathcal{O}\left(T^{-1/4}(\log T)^{1/2}\right).
\end{equation}
\end{theorem}
The detailed proof can be found in Appendix \ref{proof: theorem_convergence}. 

Observe that the bounds outlined in Theorem \ref{theorem: convergence} are expressed in terms of $\ell_2$ norms. This is different from the conventional use of squared $\ell_2$ norms prevalent in the majority of two-timescale convergent algorithms, as demonstrated in \cite{hong2023two,xu2019two,xu2021sample,ma2021greedy,zeng2022maximum,wang2022finite,wang2020finite}. The primary reason for this distinction lies in the normalization of the gradient employed in update \eqref{eq: updating_theta}, which is common in the optimization techniques like normalized or clipped gradient-based algorithms. However, it can be inferred that the norm scaling at an order of $T^{-1/4}(\log T)^{1/2}$ is roughly equivalent to the squared norm scaling at an order of $T^{-1/2}\log T$, albeit with a slight discrepancy.

\begin{remark}

The rates in \eqref{eq: theorem_bound_1} and \eqref{eq: theorem_bound_3} have components that are inversely proportional to the square of $\lambda_g$, the lower bound of the eigenvalue of $\Sigma$. This suggests that a larger $\lambda_g$ leads to faster convergence of the lower-level problem, thereby enhancing the convergence of the upper-level problem. Additionally, a high value of $\lambda_g$, along with a large coefficient $\tau$, augments the smoothness of the objective function, as established in Lemma \ref{lemma: J_smoothness}. Furthermore, the rates in \eqref{eq: theorem_bound_2} and \eqref{eq: theorem_bound_3} include components inversely proportional to $\sigma_{\min}$, the lower bound of the singular value of the matrix $\hat{\Sigma}_{\tau,\delta,\theta}$. Increased smoothness in $J$, together with a large $\sigma_{\min}$, allows for a greater step size $\alpha$. This, in practice, accelerates convergence. The effect of the threshold $\delta$ on the rates in Theorem \ref{theorem: convergence} is more complex, and we defer the discussion to Section \ref{sec: threshold}.
\end{remark}

\begin{remark}
The convergence rate of Algorithm \ref{algo} is constrained by the convergence rate \eqref{eq: theorem_bound_1} of its lower-level problem. In a nested loop algorithm framework, where the lower-level problem is optimally solved in each iteration, we have $\omega^{t+1} = \omega^*(\theta^t)$ and $\bar{\nabla}_\theta f(\theta^{t},\omega^{t+1}) = \nabla J(\theta^t)$. As a result, the bound defined in $\eqref{eq: J_grad}$ can be achieved as $\mathcal{O}(T^{-1/2}\log T)$ by choosing a step size $\alpha= \alpha_0T^{-1/2}$, where $T$ denotes the total number of iterations for updating the target network $\hat{Q}_{\theta^t}$.
\end{remark}

We also demonstrate that the target network converges to the main network.
\begin{corollary}\label{corollary: target_convergence}
The target network $\hat{Q}_{\theta^t}$ converges to the main network $\hat{Q}_{\omega^t}$ at a rate given by
\[
\frac{1}{T}\sum_{t = 0}^{T-1}\mathbb{E} \left\|\theta^t-\omega^t\right\| = \mathcal{O}\left(T^{-1/4}(\log T)^{1/2}\right).
\]
 \end{corollary} 
The detailed proof is in Appendix \ref{proof: corollary_target_convergence}.

\subsection{Performance Analysis}
In this subsection, we analyze the difference between the estimated state-action value function $ \hat{Q}_{\theta} $ and the optimal value function $ Q^*_{\tau} $ in $ \mathcal{M}_{\tau} $.

The approximation error in our linear function approximation setting is defined as
\begin{equation}\label{eq: approximation_error}
    \mathcal{E}_{\mathrm{approx}} \triangleq \sup_{\theta\in\mathbb{R}^d} \left\|\Pi\mathcal{B}_{\tau,\delta}\hat{Q}_{\theta} - \mathcal{B}_{\tau,\delta}\hat{Q}_{\theta} \right\|_\infty.
\end{equation}
This error quantifies the representational power of the linear approximation architecture \cite{bhandari2018finite}. With additional information about the problems we are targeting, the space containing $\theta$ to be considered can be narrowed. Consequently, this leads to a further reduction in the approximation error. Also, we define 
\[
\delta_{0}\triangleq\frac{R_{\max}+\tau B}{1-\gamma}, 
\]
which serves as an upper bound for all state value functions in $\mathcal{M}_{\tau}$.

In the following theorem, we present the finite-time bound characterizing the difference between the estimated value function and the optimal value functions.
\begin{theorem}\label{theorem: estimation_bound}
    Suppose Assumptions \ref{assump: Ergodicity}, \ref{assump: finite_feature} and \ref{assump: non-singular} hold. Then Algorithm \ref{algo} yields 
      \begin{align}\label{eq: estimation_bound}
        \frac{1-\gamma}{T} &\sum_{t= 0}^{T-1}  \mathbb{E} \left\|Q^*_{\tau} - \hat{Q}_{\theta^t} \right\|_{\infty} \nonumber\\ 
        \leq& \frac{1}{\sigma_{\min} T}\sum_{t= 0}^{T-1}\mathbb{E} \left\|\nabla J\left(\theta^t\right)\right\| + \mathcal{E}_{\mathrm{approx}} + \gamma\left(\delta_{0}-\mathcal{K}_{\delta}\left(\delta_{0}\right)\right) .
    \end{align} 
\end{theorem}
The detailed proof can be found in Appendix \ref{proof: theorem_estimation_bound}.

The first term in \eqref{eq: estimation_bound} can be bounded by leveraging Theorem \ref{theorem: convergence}. The term $\mathcal{E}_{\mathrm{approx}}$ is generally non-removable, a characteristic echoed in much of the literature on RL with linear function approximation \cite{tsitsiklis1997analysis,carvalho2020new,chen2022target}. The last term arises from the application of the smooth truncation operator~$\mathcal{K}_{\delta}$. We defer the discussion of this term to Section~\ref{sec: threshold}. 

As the approximation error nears zero and the effect of the smooth truncation diminishes, the learned policy $\pi_{\theta^t}$ converges towards the optimal policy $\pi^*_{\tau}$. This convergence is realized under the condition of \textit{Bellman Completeness} \cite{zanette2022realizability}, provided that the threshold $\delta$ is also sufficiently large. 

A corollary to Theorem \ref{theorem: estimation_bound} characterizes policy convergence by a straightforward application of Proposition~\ref{prop: regularizer}(iii):
\begin{corollary}\label{corollary: policy_convergence}
Suppose Assumptions \ref{assump: Ergodicity}, \ref{assump: finite_feature} and \ref{assump: non-singular} hold. Then:
\begin{align*}
        \frac{1-\gamma}{T} \!\sum_{t= 0}^{T-1}\!  \mathbb{E} \left\|\pi^*_{\tau} - \pi_{\theta^t} \right\|_{\infty} 
        &\!\leq \!\frac{\sqrt{|\mathcal{A}|}L_G}{\tau}\Big(\!\frac{1}{\sigma_{\min} T}\sum_{t= 0}^{T-1}\mathbb{E} \left\|\nabla J\left(\theta^t\right)\right\| \\
        &+ \mathcal{E}_{\mathrm{approx}} + \gamma\left(\delta_{0}-\mathcal{K}_{\delta}\left(\delta_{0}\right)\right)\Big).
\end{align*}

\end{corollary}
The detailed proof is in Appendix \ref{proof: corollary_policy_convergence}.

\section{A Discussion on Threshold $\delta$}\label{sec: threshold}
The threshold $\delta$ utilized in \cite{chen2022target} is set at $R_{\max}/(1-\gamma)$ for the hard truncation operator. This value ensures that the true optimal value function $Q^*$ is not excluded from the candidate function space under the Bellman completeness condition. 

In our context, consider a function class 
\[ 
\mathcal{Q}_{\delta} \triangleq \{\Phi\theta: \|\Phi\theta\|_{D_{\mu_{\mathrm{bhv}}}}\leq R_{\max}+\gamma \delta, \theta\in \mathbb{R}^d\}.
\]
We can ensure that $Q_{\tau}^* \in \mathcal{Q}_{\delta}$, provided that $Q_{\tau}^*$ is within the subspace of the linear span of the basis functions $\{\phi_i\}_{i=1}^d$, by choosing $\delta \geq \delta_{0}$.

We assert that the projected smooth truncated optimal regularized Bellman backup of a state-action value function $\hat{Q}$, expressed as $\Pi \mathcal{B}_{\tau,\delta}\hat{Q}$, is also an element of $\mathcal{Q}_{\delta}$ for each $\hat{Q} \in \mathcal{Q}_{\delta}$, i.e., $\Pi \mathcal{B}_{\tau,\delta}\hat{Q} \in \mathcal{Q}_{\delta}, \forall \hat{Q} \in \mathcal{Q}_{\delta}$. Given that $\mathcal{Q}_{\delta}$ is both convex and compact, the Brouwer fixed-point theorem ensures the existence of at least one fixed point $\hat{Q}^* \in \mathcal{Q}_{\delta}$ such that $\hat{Q}^* = \Pi \mathcal{B}_{\tau,\delta}\hat{Q}^*$. It is noteworthy that this assurance is absent when dealing with the composed operator $\Pi \mathcal{B}_{\tau}$ where truncation is not present. 

Unfortunately, $Q_{\tau}^*$ is not a fixed point of $\Pi \mathcal{B}_{\tau,\delta}$ even when $Q_{\tau}^* \in \mathcal{Q}_{\delta}$, except for the case where $V_{\tau}^* = \boldsymbol{0}$. Recall that the error induced by the smooth truncation operator is also accounted for in the bounds specified in Theorems \ref{theorem: estimation_bound}. To address this challenge, choosing a larger value of~$\delta$ is beneficial due to the following inequality:
\begin{equation}
\left\vert x - \mathcal{K}_{\delta_{1}} \left(x\right)\right\vert \leq \left\vert x - \mathcal{K}_{\delta_{2}} \left(x\right)\right\vert,\ \forall x\in \mathbb{R}, \delta_{1}\geq \delta_{2} \in \mathbb{R}.
\end{equation}
This suggests that by opting for a sufficiently large $\delta$, the operator $\mathcal{B}_{\tau,\delta}$ can be made to closely approximate $\mathcal{B}_{\tau}$, reducing the bias in Theorem~\ref{theorem: estimation_bound}. For instance, opting for $\delta = \delta_0$ gives us the gap $\delta_0 - \mathcal{K}_{\delta}\left(\delta_0\right) \approx 0.2384 \delta_0$. By selecting a threshold of $\delta = 10 \delta_0$, we can significantly reduce this error, as demonstrated by $\delta_0 - \mathcal{K}_{\delta}\left(\delta_0\right) \approx 0.0033 \delta_0$. 

The drawbacks of a large threshold value for $\delta$ are two-fold. Firstly, a larger value of $\delta$ increases the $\ell_2$-norms of both the stochastic gradient $h_g^t$ and the optimal point $\omega^*(\theta^t)$ in the lower-level problem. This leads to a higher variance in both the main network and the target network, affecting the convergence bounds as outlined in Theorem~\ref{theorem: convergence}. More specifically, the bound in \eqref{eq: theorem_bound_1} increases linearly, and the bound in \eqref{eq: theorem_bound_2} increases quadratically with $\delta$ when $\delta$ is large. Secondly, an increase in $\delta$ correlates with potential higher approximation error, as defined in \eqref{eq: approximation_error}. This escalation adversely affects the accuracy of the estimated value functions, potentially undermining the performance of the resulting policies in the worst case. 

A smaller $\delta$, similar to the effect of the coefficient $\tau$, reduces the smoothness of the function $J$, as detailed in Lemma 1. More importantly, the application of the smooth truncation operator $\mathcal{K}\delta$ subtly alters the discount factor $\gamma$. This modification is different from the methods used in \cite{limregq,zhang2021breaking}, which uniformly reduce the discount factor across all states. In contrast, the use of $\mathcal{K}_{\delta}$ results in a more significant impact on states $s$ with values $|G^*_{\tau}(\hat{Q}_{\theta}(s,\cdot))|$ close to $\delta$, for which $z(s;\tau,\delta,\theta)$ is small, while states with values closed to $0$ are affected to a lesser extent. However, a small $\delta$ affects more states than a large $\delta$ and consequently introduces a larger bias in the bounds stated in Theorem~\ref{theorem: estimation_bound}, as it effectively transforms the problem into one with a significantly smaller discount factor.

\section{Numerical Illustration}
In this section, we explore two environments: \textit{GridWorld} and \textit{MountainCar-v0}. We begin with the GridWorld environment, where we elucidate the influences of the smooth truncation operator $\mathcal{K}_{\delta}$ on the fixed points. Graphical illustrations complement our discussion, offering a visual representation of the convergence of the estimation from the proposed algorithm and illustrating the reduction in MSPBE. Subsequently, we assess the policies derived from our algorithm within the \textit{MountainCar-v0} environment and compare their performance with policies stemming from existing algorithms.
\subsection{GridWorld}
\begin{figure}
        \centering
        \begin{subfigure}[t]{0.3\textwidth}
            \centering
            \includegraphics[width=\textwidth]{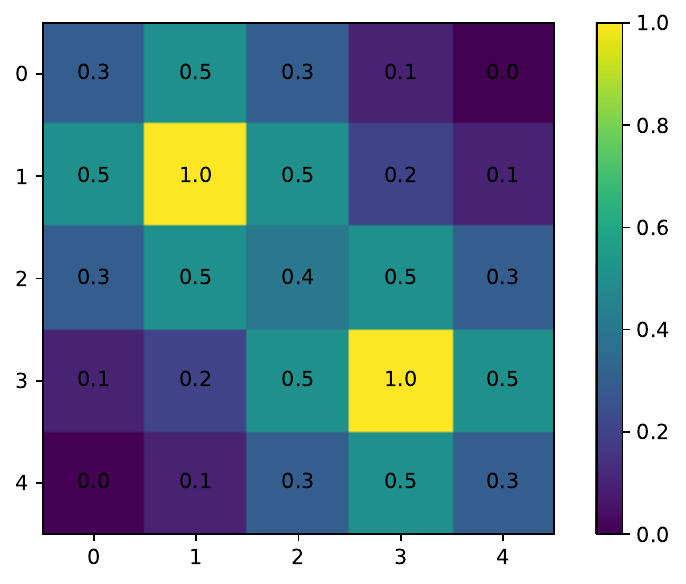} 
            \caption{Rewards Map $R$.}
            \label{fig:reward}
        \end{subfigure}
        \hfill
        \begin{subfigure}[t]{0.3\textwidth}  
            \centering 
            \includegraphics[width=\textwidth]{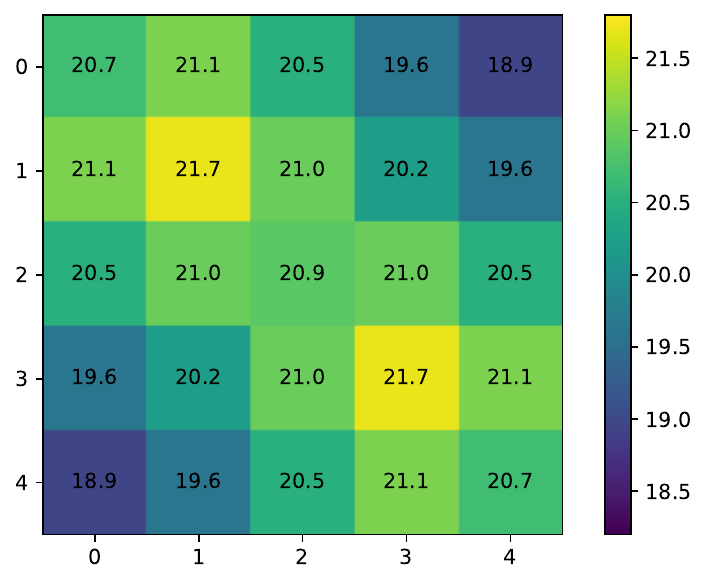}
            \caption{True Value Map $V^*_{\tau}$.}
            \label{fig:true_value}
        \end{subfigure}
        \hfill
        \begin{subfigure}[t]{0.3\textwidth}
            \centering
            \includegraphics[width=\textwidth]{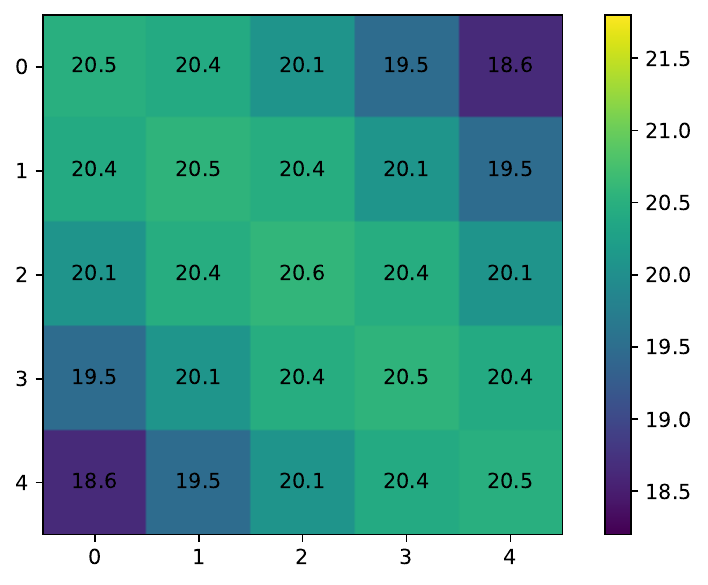}
            \caption{Fixed Point of $\Pi\mathcal{B}_{\tau}$}
            \label{fig:fixed_point}
        \end{subfigure}
        \vskip\baselineskip
        \begin{subfigure}[t]{0.3\textwidth}  
            \centering 
            \includegraphics[width=\textwidth]{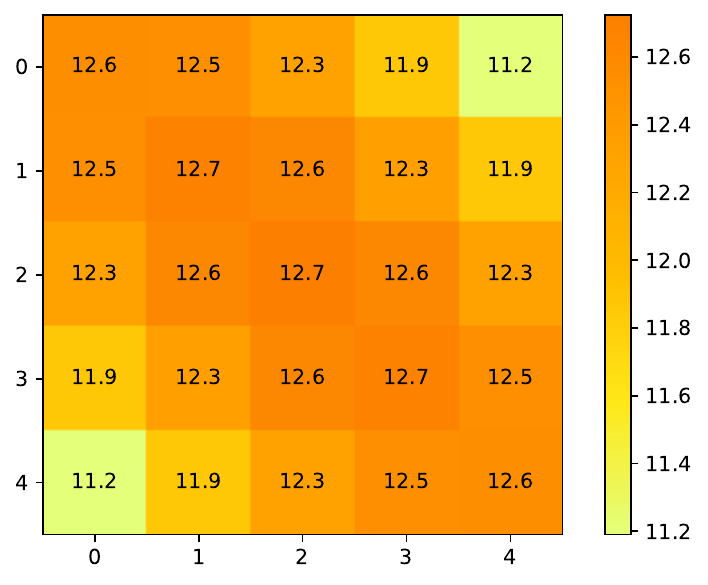}
            \caption{Fixed Point of $\Pi\mathcal{B}_{\tau,\delta_0}$}
            \label{fig:fixed_point_tr_c=1}
        \end{subfigure}
        \hfill
        \begin{subfigure}[t]{0.3\textwidth}   
            \centering 
            \includegraphics[width=\textwidth]{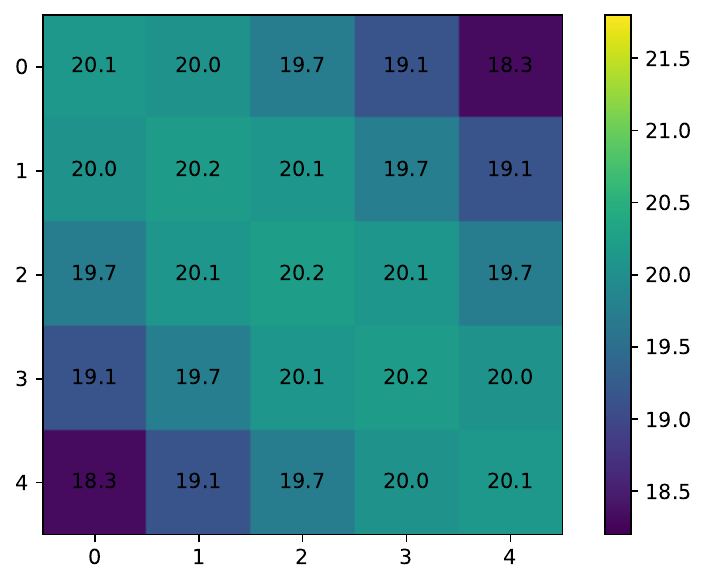}
            \caption{Fixed Point of $\Pi\mathcal{B}_{\tau,10\delta_0}$}
            \label{fig:fixed_point_tr_c=10}
        \end{subfigure}
        \hfill
        \begin{subfigure}[t]{0.3\textwidth}   
            \centering 
            \includegraphics[width=\textwidth]{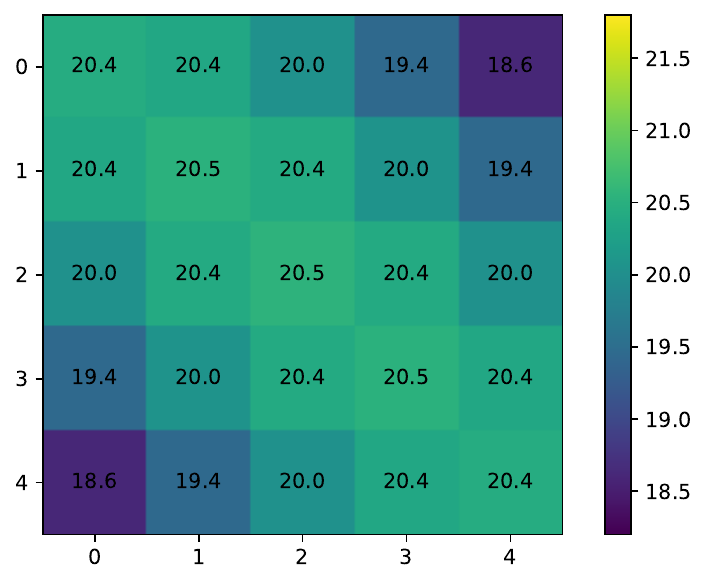}
            \caption{Fixed Point of $\Pi\mathcal{B}_{\tau,30\delta_0}$}
            \label{fig:fixed_point_tr_c=30}
        \end{subfigure}
        \caption{Rewards and Values Maps.}
        \label{fig:values}
    \end{figure}

We investigate the influence of the smooth truncation operator within a discrete GridWorld setting, characterized by a $5 \times 5$ state space and five possible actions: up, down, left, right, and stay. We consider the context of an entropy-regularized Markov Decision Process (MDP), employing a regularizer $G(\pi(\cdot \mid s)) = \langle \pi(\cdot \mid s), \log \pi(\cdot \mid s)\rangle$ with a coefficient $\tau = 1$, and a discount factor $\gamma = 0.9$.
The dynamics of agent movement are straightforward. The agent transitions between states as directed by its chosen action, except when attempting to cross the grid boundaries, where it remains at its current state.
The initial distribution is set to be uniform across all states, a setup that extends to the behavior policy as the probability of each possible action is the same. Consequently, each state-action pair in the space $\mathcal{S} \times \mathcal{A}$ has an equal likelihood of occurrence, leading to a stationary distribution $\mu_{\mathrm{bhv}}(s,a) = \frac{1}{|\mathcal{S}| \times |\mathcal{A}|} = \frac{1}{125}$ for all $(s,a)$. The reward for each state is depicted in Fig. \ref{fig:reward}.

We adopt second-order polynomial functions of the state as basis functions. Fig. \ref{fig:true_value} depicts the optimal regularized value function $V_{\tau}$. We conduct a comparison of the fixed point and the projected regularized Bellman equation, exploring scenarios both with and without the incorporation of a smooth truncation operator, all within the context of linear function approximation. 
The threshold of the smooth truncation $\mathcal{K}_{\delta}$ is defined as $\delta=c\delta_0 = \frac{c(1+\log 5)}{1-\gamma} $, with $c$ taking values of $1$, $10$, and $30$. 
From the plots, it is evident that a small threshold $(c=1)$, as shown in Fig. \ref{fig:fixed_point_tr_c=1}, introduces a significant bias into the estimation of the value function, in comparison to the fixed point obtained without applying a smooth truncation operator, depicted in Fig. \ref{fig:fixed_point}. Increasing the value of $\delta$ effectively mitigates this issue, as illustrated in Fig. \ref{fig:fixed_point_tr_c=10} and Fig. \ref{fig:fixed_point_tr_c=30}.

\begin{figure}
\centering
  \centering
  \includegraphics[width=0.6\linewidth]{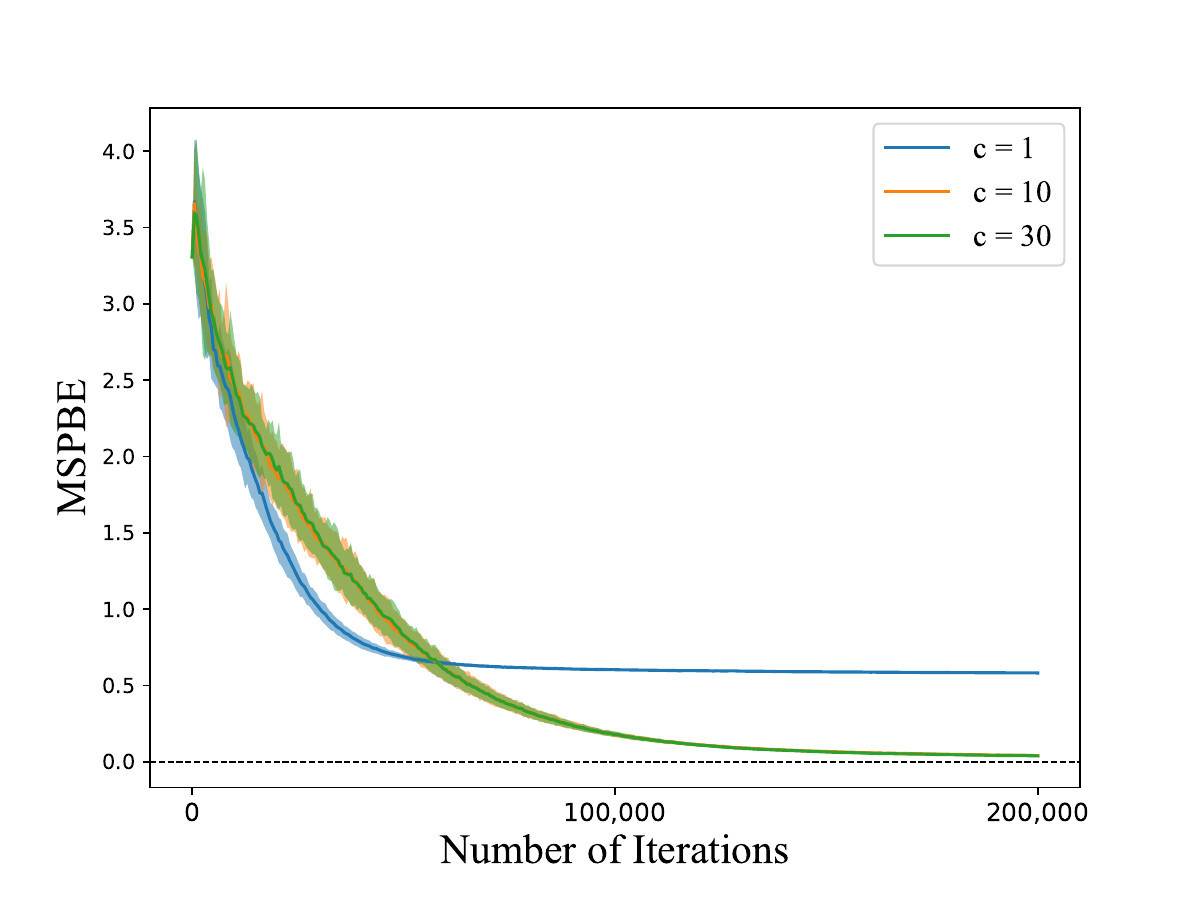}
  \caption{MSPBE of the estimated state-action value functions. The graph shows the average MSPBE ($\pm$ standard deviation) over $100$ runs.} 
  \label{fig:convergence}
\end{figure}

Next, we evaluate the convergence of Algorithm \ref{algo}. Our focus is on evaluating the Mean-Square Projected Bellman Error (MSPBE) within the regularized MDP $\mathcal{M}_{\tau}$, characterized by
\[
\operatorname{MSPBE}(\theta) \triangleq \left(\Phi\theta - \Pi \mathcal{B}_{\tau}\Phi\theta\right)^\top D_{\mu_{\mathrm{bhv}}}\left(\Phi\theta - \Pi \mathcal{B}_{\tau}\Phi\theta\right).
\]
We apply Algorithm \ref{algo} to data gathered using $\pi_{\mathrm{bhv}}$, selecting parameters $\alpha = 0.05$ and $\beta = 0.5$. Fig. \ref{fig:convergence} demonstrates that Algorithm \ref{algo} achieves convergence even with a large threshold ($c = 30$). This figure also elucidates the inherent trade-off between the convergence rate and the error induced by the smooth truncation operator. Specifically, a lower threshold facilitates quicker convergence but results in a higher MSPBE post-convergence. Conversely, increasing the threshold yields a reduction in MSPBE at the expense of a slightly slower convergence rate.

\subsection{MountainCar-v0}

\begin{table*}[t]
\centering
\resizebox{0.85\textwidth}{!}{%
\begin{tabular}{|c|c|c|c|c|c|}
\hline
d  & Q-learning                                                      & CQL & Double QL  &Algorithm \ref{algo}$,\tau = 0.01$                                   &Algorithm \ref{algo}$,\tau = 0.05$                                   \\ \hline
30 &$-177.28 \pm  32.00$ & $-199.37\pm 06.27$ & $-177.51\pm 33.73$ & $\boldsymbol{-144.36\pm 26.46}$ & $-177.83\pm 30.09$ \\ \hline
60 &$-143.08 \pm  32.47$ & $-200.00\pm 00.00$ & $-132.76\pm 31.54$ & $\boldsymbol{-121.01\pm 31.82}$ & $-123.84\pm 28.35$ \\ \hline
90 &$-175.40 \pm  17.77$ & $-200.00\pm 00.00$ & $-146.78\pm 22.74$ & $-121.37\pm 19.84$ & $\boldsymbol{-120.31\pm 19.20}$ \\ \hline
120 &$\boldsymbol{-138.74 \pm  11.00}$ & $-199.12\pm 03.97$ & $-146.23\pm 12.77$ & $-176.17\pm 22.65$ & $-145.15\pm 29.48$ \\ \hline
150 &$-141.04 \pm  24.77$ & $-200.00\pm 00.00$ & $-155.60\pm 19.43$ & $-139.58\pm 20.59$ & $\boldsymbol{-120.17\pm 20.64}$ \\ \hline
180 &$-135.62 \pm  32.01$ & $-196.32\pm 13.61$ & $\boldsymbol{-110.66\pm 20.87}$ & $-120.88\pm 22.50$ & $-122.64\pm 22.01$ \\ \hline
\end{tabular}}
\caption{Performance of Benchmark Algorithms over 20 Runs.}
\label{tb: policy_performance}
\end{table*}
The MountainCar-v0 environment, a standard control task available in OpenAI Gym \cite{brockman2016openai}, challenges an agent to maneuver a car placed between two hills to ascend the hill on the right. The car is characterized by a two-dimensional state space that encompasses its horizontal position and velocity. The agent can choose from three distinct actions at each timestep: accelerate to the left, accelerate to the right, or do nothing.
For every timestep that elapses before the car reaches the summit, the agent incurs a reward of $-1$. Consequently, the agent's objective is to reach the summit as fast as possible to maximize its cumulative reward. Given that the episode is capped at $200$ timesteps, the agent’s return is confined within the range of $[-200,0]$. 

We employ Radial Basis Functions (RBFs) as our basis functions and incorporate $l$ Gaussian kernels, each with a width of $1$, randomly selected from the state space. Since each RBF corresponds to three distinct actions, the total number of features is $d = 3l$. We set the stepsize parameters to $\alpha = 0.1$ and $\beta = 0.1$, and the threshold for Algorithm \ref{algo} is assigned a value of $\delta = 500$. Considering the environment's finite horizon, we adopt a discount factor of $\gamma = 1$.

We evaluated the policies derived from Algorithm \ref{algo} with a negative entropy regularizer and compared them with those obtained through Q-learning \cite{watkins1992q}, Coupled Q-learning (CQL) \cite{carvalho2020new}, Greedy-GQ \cite{maei2010toward}, and Double Q-learning (Double QL) \cite{hasselt2010double}. We selected the coefficient of the regularizer for Algorithm \ref{algo} from the set $\{0.01,0.05\}$. 
For Algorithm \ref{algo}, we adopted the regularized policy $\pi_{\theta}$ defined in \ref{eq: pi_def} as the behavioral policy. In contrast, an $\epsilon$-greedy policy, with $\epsilon = 0.1$, served as the behavioral policy for the other mentioned algorithms.

Table \ref{tb: policy_performance} summarizes the policy performances, with the exception of Greedy-GQ, which consistently registered a return of $-200$ across all testing episodes. Each of the $20$ training phases, differentiated by random seeds, comprised $1,000$ episodes. The table displays the average returns obtained during testing, averaged over $20$ runs, each encompassing $10$ episodes. The results demonstrate that the policies derived from Algorithm \ref{algo} outperform those obtained from other algorithms in most cases.

\bibliographystyle{IEEEtran}
\bibliography{IEEEabrv,ref}

\begin{thebibliography}{10}
\providecommand{\url}[1]{#1}
\csname url@samestyle\endcsname
\providecommand{\newblock}{\relax}
\providecommand{\bibinfo}[2]{#2}
\providecommand{\BIBentrySTDinterwordspacing}{\spaceskip=0pt\relax}
\providecommand{\BIBentryALTinterwordstretchfactor}{4}
\providecommand{\BIBentryALTinterwordspacing}{\spaceskip=\fontdimen2\font plus
\BIBentryALTinterwordstretchfactor\fontdimen3\font minus \fontdimen4\font\relax}
\providecommand{\BIBforeignlanguage}[2]{{%
\expandafter\ifx\csname l@#1\endcsname\relax
\typeout{** WARNING: IEEEtran.bst: No hyphenation pattern has been}%
\typeout{** loaded for the language `#1'. Using the pattern for}%
\typeout{** the default language instead.}%
\else
\language=\csname l@#1\endcsname
\fi
#2}}
\providecommand{\BIBdecl}{\relax}
\BIBdecl

\bibitem{Tishby_2011}
N.~Tishby and D.~Polani, \emph{Information Theory of Decisions and Actions}.\hskip 1em plus 0.5em minus 0.4em\relax New York, NY: Springer New York, 2011, pp. 601--636.

\bibitem{Ortega_2013}
P.~A. Ortega and D.~A. Braun, ``Thermodynamics as a theory of decision-making with information-processing costs,'' \emph{Proceedings of the Royal Society A: Mathematical, Physical and Engineering Sciences}, vol. 469, no. 2153, p. 20120683, 2013.

\bibitem{Matejka_2015}
F.~Matejka and A.~McKay, ``Rational inattention to discrete choices: A new foundation for the multinomial logit model,'' \emph{American Economic Review}, vol. 105, no.~1, p. 272–98, January 2015.

\bibitem{Hansen_2018}
L.~P. Hansen and J.~Miao, ``Aversion to ambiguity and model misspecification in dynamic stochastic environments,'' \emph{Proceedings of the National Academy of Sciences}, vol. 115, no.~37, pp. 9163--9168, 2018.

\bibitem{schulman2015trust}
J.~Schulman, S.~Levine, P.~Abbeel, M.~Jordan, and P.~Moritz, ``Trust region policy optimization,'' in \emph{International conference on machine learning}.\hskip 1em plus 0.5em minus 0.4em\relax PMLR, 2015, pp. 1889--1897.

\bibitem{haarnoja2018soft}
T.~Haarnoja, A.~Zhou, P.~Abbeel, and S.~Levine, ``Soft actor-critic: Off-policy maximum entropy deep reinforcement learning with a stochastic actor,'' in \emph{International conference on machine learning}.\hskip 1em plus 0.5em minus 0.4em\relax PMLR, 2018, pp. 1861--1870.

\bibitem{geist2019theory}
M.~Geist, B.~Scherrer, and O.~Pietquin, ``A theory of regularized markov decision processes,'' in \emph{International Conference on Machine Learning}.\hskip 1em plus 0.5em minus 0.4em\relax PMLR, 2019, pp. 2160--2169.

\bibitem{yang2019regularized}
W.~Yang, X.~Li, and Z.~Zhang, ``A regularized approach to sparse optimal policy in reinforcement learning,'' \emph{Advances in Neural Information Processing Systems}, vol.~32, 2019.

\bibitem{eysenbach2021maximum}
B.~Eysenbach and S.~Levine, ``Maximum entropy {RL} (provably) solves some robust {RL} problems,'' \emph{arXiv preprint arXiv:2103.06257}, 2021.

\bibitem{Haarnoja_2017}
T.~Haarnoja, H.~Tang, P.~Abbeel, and S.~Levine, ``Reinforcement learning with deep energy-based policies,'' in \emph{Proceedings of the 34th International Conference on Machine Learning - Volume 70}, ser. ICML'17.\hskip 1em plus 0.5em minus 0.4em\relax JMLR.org, 2017, p. 1352–1361.

\bibitem{lee2018sparse}
K.~Lee, S.~Choi, and S.~Oh, ``Sparse {M}arkov decision processes with causal sparse {T}sallis entropy regularization for reinforcement learning,'' \emph{IEEE Robotics and Automation Letters}, vol.~3, no.~3, pp. 1466--1473, 2018.

\bibitem{martins2016softmax}
A.~Martins and R.~Astudillo, ``From softmax to sparsemax: A sparse model of attention and multi-label classification,'' in \emph{International conference on machine learning}.\hskip 1em plus 0.5em minus 0.4em\relax PMLR, 2016, pp. 1614--1623.

\bibitem{baird1995residual}
L.~Baird, ``Residual algorithms: Reinforcement learning with function approximation,'' in \emph{Machine Learning Proceedings 1995}.\hskip 1em plus 0.5em minus 0.4em\relax Elsevier, 1995, pp. 30--37.

\bibitem{hong2023two}
M.~Hong, H.-T. Wai, Z.~Wang, and Z.~Yang, ``A two-timescale stochastic algorithm framework for bilevel optimization: Complexity analysis and application to actor-critic,'' \emph{SIAM Journal on Optimization}, vol.~33, no.~1, pp. 147--180, 2023.

\bibitem{meyn2023stability}
S.~Meyn, ``Stability of q-learning through design and optimism,'' \emph{arXiv preprint arXiv:2307.02632}, 2023.

\bibitem{sutton2018reinforcement}
R.~S. Sutton and A.~G. Barto, \emph{Reinforcement learning: An introduction}.\hskip 1em plus 0.5em minus 0.4em\relax MIT Press, 2018.

\bibitem{mnih2015human}
V.~Mnih, K.~Kavukcuoglu, D.~Silver, A.~A. Rusu, J.~Veness, M.~G. Bellemare, A.~Graves, M.~Riedmiller, A.~K. Fidjeland, G.~Ostrovski \emph{et~al.}, ``Human-level control through deep reinforcement learning,'' \emph{Nature}, vol. 518, no. 7540, pp. 529--533, 2015.

\bibitem{hasselt2010double}
H.~Hasselt, ``Double q-learning,'' \emph{Advances in neural information processing systems}, vol.~23, 2010.

\bibitem{ernst2005tree}
D.~Ernst, P.~Geurts, and L.~Wehenkel, ``Tree-based batch mode reinforcement learning,'' \emph{Journal of Machine Learning Research}, vol.~6, 2005.

\bibitem{maei2010toward}
H.~R. Maei, C.~Szepesv{\'a}ri, S.~Bhatnagar, and R.~S. Sutton, ``Toward off-policy learning control with function approximation.'' in \emph{ICML}, vol.~10, 2010, pp. 719--726.

\bibitem{chen2020zap}
S.~Chen, A.~M. Devraj, F.~Lu, A.~Busic, and S.~Meyn, ``Zap q-learning with nonlinear function approximation,'' \emph{Advances in Neural Information Processing Systems}, vol.~33, pp. 16\,879--16\,890, 2020.

\bibitem{devraj2017zap}
A.~M. Devraj and S.~Meyn, ``Zap q-learning,'' \emph{Advances in Neural Information Processing Systems}, vol.~30, 2017.

\bibitem{meyn2022control}
S.~Meyn, \emph{Control systems and reinforcement learning}.\hskip 1em plus 0.5em minus 0.4em\relax Cambridge University Press, 2022.

\bibitem{chen2022target}
Z.~Chen, J.~P. Clarke, and S.~T. Maguluri, ``Target network and truncation overcome the deadly triad in $ q $-learning,'' \emph{arXiv preprint arXiv:2203.02628}, 2022.

\bibitem{ma2021greedy}
S.~Ma, Z.~Chen, Y.~Zhou, and S.~Zou, ``{Greedy-GQ} with variance reduction: Finite-time analysis and improved complexity,'' \emph{arXiv preprint arXiv:2103.16377}, 2021.

\bibitem{carvalho2020new}
D.~Carvalho, F.~S. Melo, and P.~Santos, ``A new convergent variant of q-learning with linear function approximation,'' \emph{Advances in Neural Information Processing Systems}, vol.~33, pp. 19\,412--19\,421, 2020.

\bibitem{borkar1997stochastic}
V.~S. Borkar, ``Stochastic approximation with two time scales,'' \emph{Systems \& Control Letters}, vol.~29, no.~5, pp. 291--294, 1997.

\bibitem{limregq}
H.-D. Lim, D.~Lee \emph{et~al.}, ``{RegQ}: Convergent q-learning with linear function approximation using regularization,'' 2023.

\bibitem{zhang2021breaking}
S.~Zhang, H.~Yao, and S.~Whiteson, ``Breaking the deadly triad with a target network,'' in \emph{International Conference on Machine Learning}.\hskip 1em plus 0.5em minus 0.4em\relax PMLR, 2021, pp. 12\,621--12\,631.

\bibitem{sutton2009fast}
R.~S. Sutton, H.~R. Maei, D.~Precup, S.~Bhatnagar, D.~Silver, C.~Szepesv{\'a}ri, and E.~Wiewiora, ``Fast gradient-descent methods for temporal-difference learning with linear function approximation,'' in \emph{Proceedings of the 26th annual international conference on machine learning}, 2009, pp. 993--1000.

\bibitem{wang2020finite}
Y.~Wang and S.~Zou, ``Finite-sample analysis of {Greedy-GQ} with linear function approximation under {M}arkovian noise,'' in \emph{Conference on Uncertainty in Artificial Intelligence}.\hskip 1em plus 0.5em minus 0.4em\relax PMLR, 2020, pp. 11--20.

\bibitem{wang2022finite}
Y.~Wang, Y.~Zhou, and S.~Zou, ``Finite-time error bounds for {Greedy-GQ},'' \emph{arXiv preprint arXiv:2209.02555}, 2022.

\bibitem{xu2021sample}
T.~Xu and Y.~Liang, ``Sample complexity bounds for two timescale value-based reinforcement learning algorithms,'' in \emph{International Conference on Artificial Intelligence and Statistics}.\hskip 1em plus 0.5em minus 0.4em\relax PMLR, 2021, pp. 811--819.

\bibitem{bhandari2018finite}
J.~Bhandari, D.~Russo, and R.~Singal, ``A finite time analysis of temporal difference learning with linear function approximation,'' in \emph{Conference on learning theory}.\hskip 1em plus 0.5em minus 0.4em\relax PMLR, 2018, pp. 1691--1692.

\bibitem{shen2020asynchronous}
H.~Shen, K.~Zhang, M.~Hong, and T.~Chen, ``Asynchronous advantage actor critic: Non-asymptotic analysis and linear speedup,'' 2020.

\bibitem{chen2019performance}
Z.~Chen, S.~Zhang, T.~T. Doan, S.~T. Maguluri, and J.-P. Clarke, ``Performance of q-learning with linear function approximation: Stability and finite-time analysis,'' \emph{arXiv preprint arXiv:1905.11425}, p.~4, 2019.

\bibitem{lee2019unified}
D.~Lee and N.~He, ``A unified switching system perspective and ode analysis of q-learning algorithms,'' \emph{arXiv preprint arXiv:1912.02270}, 2019.

\bibitem{melo2008analysis}
F.~S. Melo, S.~P. Meyn, and M.~I. Ribeiro, ``An analysis of reinforcement learning with function approximation,'' in \emph{Proceedings of the 25th international conference on Machine learning}, 2008, pp. 664--671.

\bibitem{xu2019two}
T.~Xu, S.~Zou, and Y.~Liang, ``Two time-scale off-policy {TD} learning: Non-asymptotic analysis over markovian samples,'' \emph{Advances in Neural Information Processing Systems}, vol.~32, 2019.

\bibitem{xu2020non}
T.~Xu, Z.~Wang, and Y.~Liang, ``Non-asymptotic convergence analysis of two time-scale (natural) actor-critic algorithms,'' \emph{arXiv preprint arXiv:2005.03557}, 2020.

\bibitem{Wu_2020}
Y.~Wu, W.~Zhang, P.~Xu, and Q.~Gu, ``A finite-time analysis of two time-scale actor-critic methods,'' in \emph{Proceedings of the 34th International Conference on Neural Information Processing Systems}, ser. NIPS '20, 2020.

\bibitem{borkar2018concentration}
V.~S. Borkar and S.~Pattathil, ``Concentration bounds for two time scale stochastic approximation,'' in \emph{2018 56th Annual Allerton Conference on Communication, Control, and Computing (Allerton)}.\hskip 1em plus 0.5em minus 0.4em\relax IEEE, 2018, pp. 504--511.

\bibitem{Bhatnagar_2015}
S.~Bhatnagar and K.~Lakshmanan, ``Multiscale q-learning with linear function approximation,'' \emph{Discrete Event Dynamical Systems}, vol.~26, pp. 477--509, 2015.

\bibitem{zhang2019gradient}
J.~Zhang, T.~He, S.~Sra, and A.~Jadbabaie, ``Why gradient clipping accelerates training: A theoretical justification for adaptivity,'' \emph{arXiv preprint arXiv:1905.11881}, 2019.

\bibitem{zhang2020improved}
B.~Zhang, J.~Jin, C.~Fang, and L.~Wang, ``Improved analysis of clipping algorithms for non-convex optimization,'' \emph{Advances in Neural Information Processing Systems}, vol.~33, pp. 15\,511--15\,521, 2020.

\bibitem{zeng2022maximum}
S.~Zeng, C.~Li, A.~Garcia, and M.~Hong, ``Maximum-likelihood inverse reinforcement learning with finite-time guarantees,'' \emph{arXiv preprint arXiv:2210.01808}, 2022.

\bibitem{tsitsiklis1997analysis}
J.~N. Tsitsiklis and B.~Van~Roy, ``An analysis of temporal-difference learning with function approximation,'' \emph{IEEE TRANSACTIONS ON AUTOMATIC CONTROL}, vol.~42, no.~5, 1997.

\bibitem{zanette2022realizability}
A.~Zanette, ``When is realizability sufficient for off-policy reinforcement learning?'' \emph{arXiv preprint arXiv:2211.05311}, 2022.

\bibitem{brockman2016openai}
G.~Brockman, V.~Cheung, L.~Pettersson, J.~Schneider, J.~Schulman, J.~Tang, and W.~Zaremba, ``{OpenAI} gym,'' \emph{arXiv preprint arXiv:1606.01540}, 2016.

\bibitem{watkins1992q}
C.~J. Watkins and P.~Dayan, ``Q-learning,'' \emph{Machine learning}, vol.~8, pp. 279--292, 1992.

\end{thebibliography}
\newpage
\begin{appendices}
\setcounter{theorem}{0}
\renewcommand{\thetheorem}{\Alph{section}\arabic{theorem}}
\section{Auxiliary Lemmas}
\begin{lemma}\label{lemma: omega_star_lipschitz}
Under Assumptions \ref{assump: Ergodicity} and \ref{assump: finite_feature}, $\omega^*(\cdot)$ is Lipschitz with respect to $\theta\in \mathbb{R}^d$ with constant $\gamma/\lambda_g$.
\end{lemma}
\begin{proof}
    The gradient of $\omega^*(\theta)$ with respect to $\theta\in \mathbb{R}^d$ is
    \begin{IEEEeqnarray*}{rl}
        \nabla_{\theta}\omega^*(\theta)\nonumber
=\gamma\Sigma^{-1}\mathbb{E}_{\mathcal{D}}\left[z\left(s^\prime;\tau, \delta,\theta\right)\sum_{a^\prime\in\mathcal{A}}\pi_{\theta}(a^\prime\mid s^\prime)\phi(s,a) \phi(s^\prime,a^\prime)^\top\right].
    \end{IEEEeqnarray*}
    The $\ell_2$-norm of the gradient of $\omega^*(\theta)$ is bounded as $\left\|\nabla_{\theta}\omega^*(\theta)\right\|\leq 1/\lambda_g$. According to the mean value theorem, we have
\begin{IEEEeqnarray*}{rCr}
    \left\|\omega^*\left(\theta_1\right)-\omega^*\left(\theta_2\right)\right\| \leq \max_{\theta} \left\|\nabla_{\theta}\omega^*(\theta)\right\|\left\|\theta_1-\theta_2\right\|\leq \frac{1}{\lambda_g} \left\|\theta_1-\theta_2\right\|.
\end{IEEEeqnarray*}
\end{proof}

\begin{lemma}\label{lemma: pi_lipschitz}
    Under Assumption \ref{assump: finite_feature}, the policy $\pi_{\theta}(\cdot \mid s)$ is Lipschitz in $\theta \in \mathbb{R}^d$ with constant $L_{G}\sqrt{|\mathcal{A}|}/\tau$ for $s\in \mathcal{S}$.
\end{lemma}
\begin{proof}
    By the properties of the regularizer and Assumption \ref{assump: finite_feature}, for $s\in \mathcal{S}$ and $\theta_1,\theta_2\in \mathbb{R}^d$, we have 
    \begin{IEEEeqnarray*}{rlr}
        \left\|\pi_{\theta_1}(\cdot \!\mid \!s)- \pi_{\theta_2}(\cdot \!\mid\! s)\right\| =& \left\|\nabla G^*_{\tau}(\hat{Q}_{\theta_1}\left(s,\cdot\right))-\nabla G^*_{\tau}(\hat{Q}_{\theta_2}\left(s,\cdot\right))  \right\|\\
        \leq& \frac{L_G}{\tau} \left\|\hat{Q}_{\theta_1}(s,\cdot)-\hat{Q}_{\theta_2}(s,\cdot)\right\|\\
    \leq & \frac{L_G\sqrt{|\mathcal{A}|}}{\tau}  \left\|\theta_1-\theta_2\right\|.
    \end{IEEEeqnarray*}
\end{proof}

\begin{lemma}\label{lemma: Sigma_hat_lipcshitz}
    Under Assumption \ref{assump: finite_feature}, the matrix $\hat{\Sigma}_{\tau,\delta,\theta}$ is Lipschitz in $\theta\in \mathbb{R}^d$ with constant $(L_{G}|\mathcal{A}|/\tau + 2/\delta)$.
\end{lemma}

\begin{proof}
We first derive the Lipschitz property for $ \hat{G}_{\tau,\delta}(\hat{Q}_{\theta}(s^\prime,\cdot))^2$
with respect to $ \theta$. The derivative of the expression with respect to $\theta$ is given by
{\small
\begin{align*}
    \nabla_{\theta} \hat{G}^2_{\tau,\delta} (\hat{Q}_{\theta}(s^\prime,\cdot)) &=\nabla_{\theta}\mathcal{K}^2_{\delta}( G^*_{\tau}(\hat{Q}_{\theta}(s^\prime,\cdot)))\\ 
    &=2\hat{G}_{\tau,\delta}(\hat{Q}_{\theta}(s^\prime,\cdot))z\left(s^\prime;\delta,\theta\right)\sum_{a\in \mathcal{A}}\pi_{\theta}(a\mid s) \phi(s,a).
\end{align*}
}
The $\ell_2$-norm of this derivative is bounded by $2\delta$. Thus, by the mean value theorem, for $s\in \mathcal{S}$ and $\theta_1,\theta_2\in \mathbb{R}^d$, we obtain
\begin{equation}\label{eq: K_square_lipschitz}
   \left\|\hat{G}^2_{\tau,\delta}(\hat{Q}_{\theta_1}(s^\prime,\cdot)) -\hat{G}^2_{\tau,\delta}(\hat{Q}_{\theta_2}(s^\prime,\cdot))\right\| \leq 2\delta \left\|\theta_1-\theta_2\right\|. 
\end{equation}
Next, we establish the Lipschitz property for $\hat{\Sigma}_{\tau,\delta,\theta}$. Given the definition of $\hat{\Sigma}_{\tau,\delta,\theta}$ in \eqref{eq: Sigma_hat}, for $\theta_1,\theta_2\in \mathbb{R}^d$, we have
\begin{align*}
\|\hat{\Sigma}_{\tau,\delta,\theta_1} -  \hat{\Sigma}_{\tau,\delta,\theta_2}\|
    &\leq   \left\|\mathbb{E}_{\mathcal{D}}\left[\phi\left(s^\prime,\cdot\right)\left(\pi_{\theta_1}\left(\cdot \mid s^\prime\right)-\pi_{\theta_2}\left(\cdot \mid s^\prime\right)\right)\phi(s,a)^\top\right]\right\| \\
    & \ \ \ + \frac{1}{\delta^2}\left\|\mathbb{E}_{\mathcal{D}}\left[\hat{G}^2_{\tau,\delta}(\hat{Q}_{\theta_2}(s^\prime,\cdot)) - \hat{G}^2_{\tau,\delta}(\hat{Q}_{\theta_1}(s^\prime,\cdot))\right]\right\|\\
    &\leq \left(\frac{ L_{G}|\mathcal{A}|}{\tau}+ \frac{2}{\delta}\right)\left\|\theta_1-\theta_2\right\|,
\end{align*}
where the last inequality is due to Lemma~\ref{lemma: pi_lipschitz} and \eqref{eq: K_square_lipschitz}.
\end{proof}

\begin{lemma}\label{lemma: theta_update_bound}
    Under Assumption \ref{assump: finite_feature}, two consecutive iterates $\theta^t$ and $\theta^{t+1}$ in Algorithm \ref{algo} for $t = 0,1,\dots,T-1$ satisfy $\|\theta^{t+1}-\theta^t\|\leq 2\alpha$.
\end{lemma}

\begin{proof}
The proof for the case of $\omega^{t+1} = \theta^t$ is trivial, since $\theta^{t+1} = \theta^t$. On the other hand, when $\omega^{t+1} \ne \theta^t$, 
\begin{align*}
    \left\|\theta^{t+1}-\theta^t\right\|
    \leq \alpha \left\|\gamma z\left(s^\prime; \tau,\delta,\theta^t\right) \sum_{a^\prime\in\mathcal{A}} \!\!\pi_{\theta^t}\left(a^\prime\mid s^\prime_t\right)\phi(s_t^\prime,a^\prime)-\phi(s_t,a_t)\right\|
    \leq  2\alpha. 
\end{align*}
\end{proof}

\begin{lemma}\label{lemma: gradient_bound}
    Under Assumption \ref{assump: finite_feature}, for $\omega,\theta \in \mathbb{R}^d$, the difference between the gradient and the surrogate gradient of $J(\cdot)$ is bounded: $\left\|\nabla J(\theta)- \bar{\nabla}_{\theta}f(\theta,\omega)\right\|\leq 2\left\|\omega^*(\theta)-\omega\right\|$.
\end{lemma}

\begin{proof}
    Given the definitions of the gradient $\nabla J(\theta)$ and the surrogate gradient $\bar{\nabla}_{\theta} f(\theta,\omega)$ in \eqref{eq: J_grad} and \eqref{eq:f_grad}, respectively, we have
    \begin{align*}
    \|\nabla J(\theta) - \bar{\nabla}_{\theta} f(\theta,\omega)\| &\leq \left\|\hat{\Sigma}_{\tau,\delta,\theta} \left(\omega^*\left(\theta\right)-\omega\right)\right\|\\
    &\leq  \left\|\hat{\Sigma}_{\tau,\delta,\theta} \right\| \, \left\| \omega^*\left(\theta\right)-\omega\right\| \\
    &\leq 2\left\| \omega^*\left(\theta\right)-\omega\right\|. 
    \end{align*}
\end{proof}
\section{Proofs of Theorems}
\setcounter{theorem}{0}
\subsection{Proof of Theorem \ref{theorem: convergence}}\label{proof: theorem_convergence}
Let $\mathcal{F}_t:=\sigma\left\{\omega^0, \theta^0, \ldots, \omega^t, \theta^t\right\}$ and $\mathcal{F}^\prime_t:=\sigma\left\{\omega^0, \theta^0, \ldots, \omega^t, \theta^t,\omega^{t+1}\right\}$ be the filtration of the random variables up to iteration $t$, where $\sigma\{\cdot\}$ denotes the $\sigma$-algebra generated by the random variables. 
Under Assumption \ref{assump: Ergodicity}, we define $T_\alpha$ as the number of iterations needed to ensure the bias incurred by the Markovian sampling is sufficiently small:
\begin{equation}\label{eq: T_alpha}
    T_\alpha \triangleq \min \left\{t \in \mathbb{N}^{+} \mid \kappa \rho^{t} \leq \alpha\right\}.
\end{equation}
We define $T_\beta$ in a similar way:
\begin{equation}\label{eq: T_beta}
    T_\beta \triangleq \min \left\{t \in \mathbb{N}^{+} \mid \kappa \rho^{t} \leq \beta\right\}.
\end{equation}

We first start with the error in the lower-level problem. The following lemma demonstrates the tracking error of the estimated main network $\hat{Q}_{\omega^{t}}$.

\begin{lemma}\label{lemma: tracking_error}
    (Tracking error). Given the conditions stated in Theorem \ref{theorem: convergence}, 
    \[
    \frac{1}{T}\sum_{t=0}^{T-1}\mathbb{E} \left\|\omega^{*}(\theta^{t})-\omega^{t+1}\right\|^2 = \mathcal{O}\left(T^{-1/2}\log T\right).
    \]
\end{lemma}
See Appendix \ref{proof: lemma_tracking_error} for a detailed proof.
As a result, by Jensen's inequality, we can derive
\begin{align}\label{eq: omega_bound_final}
    \frac{1}{T}\sum_{t=0}^{T-1}\mathbb{E}\left\|\omega^{*}(\theta^{t})-\omega^{t+1}\right\| 
    \leq  \sqrt{\frac{1}{T}\sum_{t=0}^{T-1}\mathbb{E}\left\|\omega^{*}(\theta^{t})-\omega^{t+1}\right\|^2}
    = \mathcal{O}\left(T^{-1/4}(\log T)^{1/2}\right).
\end{align}

Now, we provide the proof for the second bound in Theorem \ref{theorem: convergence}. Toward this end, we use the following lemma. 
\begin{lemma}\label{lemma: upper_error}
    Given the conditions stated in Theorem \ref{theorem: convergence},
    \[
     \frac{1}{T}\sum_{t= 0}^{T-1} \left\|\bar{\nabla}_\theta f\left(\theta^t,\omega^{t+1}\right) \right\|
    = \mathcal{O}\left(T^{-1/4}(\log T)^{1/2}\right) .
    \]
\end{lemma}
See Appendix \ref{proof: lemma_upper_error} for a detailed proof.

Finally, it is straightforward to derive~\eqref{eq: theorem_bound_3}. By the triangle inequality and Lemma \ref{lemma: gradient_bound}, we have
\begin{align*}
    \frac{1}{T}\sum_{t= 0}^{T-1} \left\|{\nabla}J\left(\theta^t\right) \right\|
    &\leq \frac{1}{T}\sum_{t= 0}^{T-1}\left(\left\|{\nabla}J\left(\theta^t\right) - \bar{\nabla}_\theta f\left(\theta^t,\omega^{t+1}\right) \right\|  + \left\|\bar{\nabla}_\theta f\left(\theta^t,\omega^{t+1}\right)\right\|\right)\nonumber\\
    &\leq  \frac{2}{T}\sum_{t= 0}^{T-1}\left\|\omega^*\left(\theta^t\right)-\omega^{t+1}\right\|
    +  \frac{1}{T}\sum_{t= 0}^{T-1}\left\|\bar{\nabla}_\theta f\left(\theta^t,\omega^{t+1}\right)\right\|\nonumber\\
    &= \mathcal{O}\left(T^{-1/4}(\log T)^{1/2}\right).
\end{align*}
The proof is concluded.

\subsection{Proof of Theorem \ref{theorem: estimation_bound}}\label{proof: theorem_estimation_bound}
We start with the first bound in Theorem \ref{theorem: estimation_bound}. For $t> 0$, the triangle inequality yields
\begin{align}\label{eq: estimation_Q_diff}
    \|Q^*_{\tau}-\hat{Q}_{{\theta^t}}\|_{\infty} \leq & \|Q^*_{\tau}-\mathcal{B}_{\tau,\delta}\hat{Q}_{{\theta^t}}\|_{\infty} \nonumber \\
    & \quad + \|\mathcal{B}_{\tau,\delta}\hat{Q}_{{\theta^t}}-\Pi\mathcal{B}_{\tau,\delta}\hat{Q}_{{\theta^t}}\|_{\infty} \nonumber\\
        & \quad \quad + \|\Pi\mathcal{B}_{\tau,\delta}\hat{Q}_{{\theta^t}}-\hat{Q}_{{\theta^t}} \|_{\infty}.
\end{align}
For the first term in \eqref{eq: estimation_Q_diff}, due to regularized Bellman equation and the $\gamma$-contraction property of the smooth truncated optimal regularized Bellman operator $\mathcal{B}_{\tau,\delta}$ in $\ell_\infty$-norm, we obtain 
\begin{align}\label{eq: estimation_Q_diff_term1}
    \|Q^*_{\tau}- \mathcal{B}_{\tau,\delta}\hat{Q}_{{\theta^t}} \|_{\infty} \!
    \leq & \|\mathcal{B}_{\tau}Q^*_{\tau}-\mathcal{B}_{\tau,\delta}Q^*_{\tau}\| + \gamma\|Q^*_{\tau}-\hat{Q}_{{\theta^t}}\|_{\infty}\nonumber\\
    =  & \gamma\max_{s\in\mathcal{S}}\left\vert V_{\tau}^* - \mathcal{K}_{\delta}\left(V_{\tau}^*\right) \right\vert+ \gamma\|Q^*_{\tau}-\hat{Q}_{{\theta^t}}\|_{\infty}\nonumber\\
    \leq & \gamma\left(\delta_{0}-\mathcal{K}_{\delta}\!\left(\delta_{0}\right)\right) + \gamma\|Q^*_{\tau}-\hat{Q}_{{\theta^t}}\|_{\infty},
\end{align}
where the last inequality follows from the fact $\|V_{\pi,\tau}\|_{\infty} \leq \frac{R_{\max}+\tau B}{1-\gamma} = \delta_0$ for any policy $\pi$ and~\eqref{eq: truncation_gap_inequality}. Next, we can bound the second term in \eqref{eq: estimation_Q_diff} by the definition of approximation error in \eqref{eq: approximation_error}:
\begin{equation}\label{eq: estimation_Q_diff_term2}
\left\|\mathcal{B}_{\tau,\delta}\hat{Q}_{{\theta^t}}-\Pi\mathcal{B}_{\tau,\delta}\hat{Q}_{{\theta^t}}\right\|_{\infty} \leq \mathcal{E}_{\mathrm{approx}}.
\end{equation}
Finally, we bound the last term in \eqref{eq: estimation_Q_diff}. Under Assumption \ref{assump: non-singular} and given the expression of $\nabla J(\cdot)$, we have
\begin{equation*}
    \left\|\theta^t-\omega^*\left(\theta^t\right)\right\| \leq \frac{1}{\sigma_{\min}}\left\|\nabla J\left(\theta^t\right)\right\|
\end{equation*}
Then, provided the equation $\hat{Q}_{\omega^*\left(\theta\right)} = \Pi \mathcal{B}_{\tau,\delta} \hat{Q}_{\theta}$, we obtain
\begin{align}\label{eq: estimation_Q_diff_term3}
\left\|\Pi\mathcal{B}_{\tau,\delta}\hat{Q}_{{\theta^t}}-\hat{Q}_{{\theta^t}}\right\|_{\infty} =& \left\|\hat{Q}_{\omega^*\left(\theta^t\right)}-\hat{Q}_{{\theta^t}}\right\|_{\infty}\nonumber\\
=& \left\|\Phi\omega^*\left(\theta^t\right)-\Phi\theta^t\right\|_{\infty}\nonumber\\
=& \max_{(s,a)\in\mathcal{S}\times\mathcal{A}}\left\vert \phi(s,a)^\top\left( \omega^*\left(\theta^{t}\right) -\theta^t \right)\right\vert\nonumber\\
\leq &\left\|\omega^*\left(\theta^{t}\right) -\theta^t\right\|\nonumber\\
\leq&  \frac{1}{\sigma_{\min}}\left\|\nabla J\left(\theta^t\right)\right\|.
\end{align}
Substituting \eqref{eq: estimation_Q_diff_term1}, \eqref{eq: estimation_Q_diff_term2} and \eqref{eq: estimation_Q_diff_term3} into \eqref{eq: estimation_Q_diff} leads to
\begin{align*}
   (1-\gamma) & \|Q^*_{\tau}-\hat{Q}_{{\theta^t}}\|_{\infty}\nonumber\\
        \leq&  \frac{1}{\sigma_{\min}} \|\nabla J\left(\theta^t\right) \|+ \gamma\left(\delta_{0}-\mathcal{K}_{\delta}\left(\delta_{0}\right)\right) + \mathcal{E}_{\mathrm{approx}}.
\end{align*}
Taking full expectation and summing from $t=0$ to $t= T-1$ for the above inequality completes the proof of~\eqref{eq: estimation_bound}. 
For the second bound in Theorem \ref{theorem: estimation_bound}, given the definition of $V_{\tau}^t$, we have
\begin{align}\label{eq: performance_bound}
     \mathbb{E} \|V^*_{\tau} - & V_{\pi_{\theta^t},\tau} \|_{\infty} \nonumber\\
    \leq &\mathbb{E}\left[\max_{s\in \mathcal{S}}\left\vert V^*_{\tau}(s) -  G^*_{\tau}(\hat{Q}_{\theta^t}\left(s,\cdot\right))\right\vert\right]\nonumber\\
    &+ \mathbb{E}\left[\max_{s\in \mathcal{S}}\left\vert G^*_{\tau}(\hat{Q}_{\theta^t}\left(s,\cdot\right))-V_{\pi_{\theta^t},\tau}(s)\right\vert\right].
\end{align}

For the first term in \eqref{eq: performance_bound}, we utilize the properties of the convex conjugate $G^*$ of the regularizer:
\begin{align}\label{eq: performance_bound_term1}
    \max_{s\in \mathcal{S}} &\left\vert V^*_{\tau}(s) - G^*_\tau\left(\hat{Q}_{\theta^t}\left(s,\cdot\right)\right)\right\vert \nonumber\\
    &= \max_{s\in \mathcal{S}}\left\vert G^*_\tau\left({Q}^*_{G,
    \tau}\left(s,\cdot\right)\right) - G^*_{\tau}\left(\hat{Q}_{\theta^t}\left(s,\cdot\right)\right)\right\vert\nonumber\\
    &\leq   \max_{s\in\mathcal{S}} \max_{q\in \mathbb{R^{|\mathcal{A}|}}} \left|\nabla G^*(q)^\top\left(Q^*_{\tau}(s,\cdot)-\hat{Q}_{{\theta^t}}(s,\cdot)\right)\right|\nonumber\\
        &\leq  \max_{s\in\mathcal{S}} \max_{a\in \mathcal{A}} \left|Q^*_{\tau}(s,a)-\hat{Q}_{{\theta^t}}(s,a)\right|\nonumber\\
        &=  \left\| Q^*_{\tau} -\hat{Q}_{{\theta^t}}\right\|_{\infty},
\end{align}
where the first inequality follows from Taylor's theorem and the second inequality is due to Proposition~\ref{prop: regularizer}. 

For the second term in \eqref{eq: performance_bound}, the bound can be developed similarly. Given definition of $\pi_{\theta}$ in \eqref{eq: pi_def}, for $s\in\mathcal{S}$, we have
\begin{align*}
    G^*_{\tau}(\hat{Q}_{\theta^t}\left(s,\cdot\right)) 
    =&  \max_{p\in \Delta(\mathcal{A})}\sum_{a}p(a)\, \hat{Q}_{{\theta^t}}(s,a) - G_{\tau} \left(p\right)\nonumber\\
     =& \sum_{a}\pi_{\theta^t}(a\mid s) \, \hat{Q}_{{\theta^t}}(s,a) - G_{\tau} \left(\pi_{\theta^t}(\cdot\mid s)\right).
\end{align*}
Therefore, by the definition of $V_{\pi,\tau}$ and $Q_{\pi,\tau}$, we can derive the bound 
\begin{align}\label{eq: performance_bound_term2}
    \max_{s\in \mathcal{S}} &\left\vert G^*_\tau(\hat{Q}_{\theta^t}\left(s,\cdot\right))-V_{\pi_{\theta^t},\tau}(s)\right\vert\nonumber\\
    &= \max_{s\in \mathcal{S}}\Bigg\vert\sum_{a}\pi_{\theta^t}(a\mid s)\hat{Q}_{{\theta^t}}(s,a)
    -  G_{\tau} \left(\pi_{\theta^t}(\cdot\mid s)\right)\nonumber\\
    & \quad -\left(\sum_{a}\pi_{\theta^t}(a\mid s)Q_{\pi_{\theta^t},\tau}(s,a) -  G_{\tau} \left(\pi_{\theta^t}(\cdot\mid s)\right)\right)\Bigg\vert\nonumber\\
    &\leq \left\|\hat{Q}_{{\theta^t}}- Q_{\pi_{\theta^t},\tau}\right\|_{\infty} \nonumber\\
    &\leq \left\|Q_{\pi_{\theta^t},\tau} - \mathcal{B}_{\tau,\delta}\hat{Q}_{\theta^t}\right\|_{\infty} + \left\|\mathcal{B}_{\tau,\delta}\hat{Q}_{\theta^t}-\Pi \mathcal{B}_{\tau,\delta}\hat{Q}_{\theta^t}\right\|_{\infty}\nonumber\\
    &\quad + \left\|\Pi \mathcal{B}_{\tau,\delta}\hat{Q}_{\theta^t}-\hat{Q}_{\theta^t}\right\|_{\infty}.
\end{align}
Here, our focus is on providing a bound for the first term, as bounds for the second and third terms have already been established in \eqref{eq: estimation_Q_diff_term2} and \eqref{eq: estimation_Q_diff_term3}. Given the regularized Bellman equation, we can derive
\begin{align}\label{eq: performance_bound_term2_term1}
    \| & Q_{\pi_{\theta^t},\tau} - \mathcal{B}_{\tau,\delta}\hat{Q}_{\theta^t} \|_{\infty} \nonumber\\
    &=\gamma \max_{s,a}\left\vert  \mathbb{E}_{s^\prime\sim P(\cdot \mid s,a )}\left[{V}_{\pi_{\theta^t},\tau}\left(s^\prime\right) - \hat{G}_{\tau,\delta}\left( \hat{Q}_{\theta^t}\left(s^\prime,\cdot\right)\right) \right]\right\vert\nonumber\\
    &\leq  \gamma \max_{s,a}\Big\vert  \mathbb{E}_{s^\prime\sim P(\cdot \mid s,a )}\Big[{V}_{\pi_{\theta^t},\tau}\left(s^\prime\right) - \mathcal{K}_{\delta}\left({V}_{\pi_{\theta^t},\tau}\left(s^\prime\right)\right)\nonumber\\
    & \ \ \ + \mathcal{K}_{\delta}\left({V}_{\pi_{\theta^t},\tau}\left(s^\prime\right)\right)-\hat{G}_{\tau,\delta}\left( \hat{Q}_{\theta^t}\left(s^\prime,\cdot\right)\right)\Big]\Big\vert\nonumber\\
     &\leq  \gamma \left(\delta_{0} - \mathcal{K}_{\delta}\left(\delta_{0}\right)\right) \nonumber\\
     & \ \ \ + 
    \gamma \max_{s\in\mathcal{S}}\left\vert{V}_{\pi_{\theta^t},\tau}\left(s\right) - G^*_\tau\left( \hat{Q}_{\theta^t}\left(s,\cdot\right)\right)\right\vert,
\end{align}
where the last inequality follows from the $1$-Lipschitz property of $\mathcal{K}_{\delta}$. Substituting \eqref{eq: performance_bound_term2_term1}, \eqref{eq: estimation_Q_diff_term2} and \eqref{eq: estimation_Q_diff_term3} into \eqref{eq: performance_bound_term2} leads to 
\begin{align*}
    \left(1-\gamma\right) &\max_{s\in \mathcal{S}} \left\vert G^*_\tau(\hat{Q}_{\theta^t}\left(s,\cdot\right))-V_{\pi_{\theta^t},\tau}(s)\right\vert \\  &\leq \frac{1}{\sigma_{\min}}\left\|\nabla J\left(\theta^t\right)\right\| +\mathcal{E}_{\mathrm{approx}}+\gamma\left(\delta_{0} - \mathcal{K}_{\delta}\left(\delta_{0}\right)\right).
\end{align*}
Combining the above inequality with \eqref{eq: performance_bound_term1} and using the result in \eqref{eq: estimation_bound} complete the proof of Theorem~\ref{theorem: estimation_bound}.
\section{Proofs of Other Lemmas}
{\subsection{Proof of Lemma \ref{lemma: stability}}\label{proof: lemma_stability}
With $\delta=\delta_0$, we now show $\sup_t \max_{a\in\mathcal{A}} \big\{|\phi(s_{t+1},a)^{\top}\theta^t|\big\}< \infty$ w.p. 1.
By contradiction, we consider two cases separately:
\begin{itemize}
    \item[a.] $\max_{a\in\mathcal{A}}\big\{|\phi(s_{t+1},a)^{\top}\theta^t|\big\}> \delta_0 + \tau B$ {\em eventually}.
    \item[b.]$\max_{a\in\mathcal{A}}\big\{|\phi(s_{t+1},a)^{\top}\theta^t|\big\}< \delta_0 + \tau B$ {\em infinitely often}.
\end{itemize}
Let us consider case (a), i.e. there is a $T<\infty$ such that
$\max_{a\in\mathcal{A}}\big\{|\phi(s_{t+1},a)^{\top}\theta^t|\big\}>\delta_0 +\tau B$ for all $t\geq T$. 
By Proposition 2.1.(i), we have $|G^*_{\tau}(\hat{Q}_{\theta^t}(s_{t+1},\cdot))|\geq \delta_0$, hence, truncation is implemented for all $t\geq T$ and:
\begin{align*}
h_g^t&= \phi(s_t,a_t)\left(\phi(s_t,a_t)^\top\omega^t - R(s_t,a_t) -\gamma \delta_0\right)
\end{align*}
for all $t\geq T$. By ergodicity assumption,
$\omega^t \rightarrow \omega^*$ and $\theta^t \rightarrow \omega^*$ where
\begin{align*}
\omega^* & \triangleq \Sigma^{-1}\mathbb{E}_{\mathcal{D}}\left[\phi(s,a)\Big(R(s,a) + \gamma \delta_0\Big)\right]
\end{align*}
However, $|\hat{G}_{\tau,\delta} \big(\hat{Q}_{\omega^*}(s_{t+1},\cdot ))\big)|<\delta_0$. A contradiction. 
For case (b), let $\{\theta^{t(n)}\}$ be a subsequence such that $\max_{a\in\mathcal{A}}\big\{|\phi(s_{t+1},a)^{\top}\theta^{t(n)}|\big\}\rightarrow \infty$.
Let $\tau(n)$ be defined as
\begin{align*}
\tau(n) &\triangleq  \max\big\{t\leq t(n): \max_{a \in \mathcal{A}}|\phi(s_{t+1},a)^{\top}\theta^t|\leq \delta_0 +\tau B\big\}
\end{align*}
By Lemma A.4 in the paper, for all $t>0$, $\|\theta^{t+1}-\theta^t\| \leq 2\alpha $. Hence, 
\begin{align*}
t(n)-\tau(n) \geq \frac{
\max_{a \in \mathcal{A}}|\phi(s_{t+1},a)^{\top}\theta^t|-(\delta_0+\tau B)}{2\alpha}   
\end{align*} By hypothesis $\max_{a\in\mathcal{A}}\big\{|\phi(s_{t+1},a)^{\top}\theta^{t(n)}|\big\}\rightarrow \infty$, hence  $t(n)-\tau(n) \rightarrow \infty$. As in the proof of case (a), 
\begin{align*}
h_g^t&= \phi(s_t,a_t)\left(\phi(s_t,a_t)^\top\omega^t - R(s_t,a_t) -\gamma \delta_0\right) ~~~~~~t \in (\tau(n),t(n)]
\end{align*}
and $\lim_n\omega^{t(n)} = \omega^*$ and $\lim_n \theta^{t(n)}=\omega^*$. A contradiction.}

\subsection{Proof of Lemma \ref{lemma: J_smoothness}}\label{proof: lemma_J_smoothness}
We first prove the first inequality in Lemma \ref{lemma: J_smoothness}. By the expression of $\nabla J$ in \eqref{eq: J_grad}, for $\theta_1,\theta_2\in \mathbb{R}^d$, we have
\begin{align*}
    \| \nabla & J\left(\theta_1\right)-\nabla J\left(\theta_2\right) \|\\
    = & \left\|\hat{\Sigma}_{\tau,\delta,\theta_1}\left(\omega^*\left(\theta_1\right)-\theta_1\right)-\hat{\Sigma}_{\tau,\delta,\theta_2}\left(\omega^*\left(\theta_2\right)-\theta_2\right)\right\|\\
    \leq & \left\|\hat{\Sigma}_{\tau,\delta,\theta_1}\right\|\left\|\omega^*\left(\theta_1\right)-\theta_1 - \omega^*\left(\theta_2\right)+\theta_2\right\|\\
    &+ \left\|\hat{\Sigma}_{\tau,\delta,\theta_1}-\hat{\Sigma}_{\tau,\delta,\theta_2}\right\|\left\|\omega^*\left(\theta_2\right)-\theta_2\right\|\\
    \leq &\left\|\hat{\Sigma}_{\tau,\delta,\theta_1}\right\|\left(\left\|\omega^*\left(\theta_1\right) - \omega^*\left(\theta_2\right)\right\|+\left\|\theta_1 - \theta_2\right\|\right)\\
    &+ \left\|\hat{\Sigma}_{\tau,\delta,\theta_1}-\hat{\Sigma}_{\tau,\delta,\theta_2}\right\|\left\|\omega^*\left(\theta_2\right)-\theta_2\right\|\\
    \leq& \frac{2}{\lambda_g}\left\|\hat{\Sigma}_{\tau,\delta,\theta_1}\right\|\left\|\theta_1 - \theta_2\right\|\\
    &+
    \left(\frac{ L_{G}|\mathcal{A}|}{\tau}+ \frac{2}{\delta}\right)\left\|\omega^*\left(\theta_2\right)-\theta_2\right\|\left\|\theta_1 - \theta_2\right\|\\
    \leq & \left(\frac{4}{\lambda_g} + \left(\frac{ L_{G}|\mathcal{A}|}{\tau}+ \frac{2}{\delta}\right)\left\|\omega^*\left(\theta_2\right)-\theta_2\right\|\right)\left\|\theta_1 - \theta_2\right\|,
\end{align*}
where the third inequality is due to Lemma~\ref{lemma: omega_star_lipschitz} and Lemma~\ref{lemma: Sigma_hat_lipcshitz}, and the last inequality follows from the definition of $\hat{\Sigma}_{\tau,\delta,\theta}$ in \eqref{eq: Sigma_hat}.

For the second inequality in Lemma \ref{lemma: J_smoothness}, we utilize the integral remainder term from the Taylor expansion and the preceding inequality:
\begin{align*}
    J\left(\theta_1\right) &- J\left(\theta_2\right) - \left\langle \nabla J\left(\theta_2\right), \theta_1-\theta_2\right\rangle \nonumber\\
    = & \int_0^1 \left\langle \nabla J\left(\xi \theta_1 + \left(1-\xi\right)\theta_2\right) - \nabla J(\theta_2), \theta_1 -\theta_2\right\rangle d \xi\nonumber\\
    \leq& \int_{0}^1 \left\|\nabla J\left(\xi \theta_1 + \left(1-\xi\right)\theta_2\right) - \nabla J(\theta_2)\right\|d\xi \cdot \left\|\theta_1-\theta_2\right\| \nonumber\\
    \leq & \int_{0}^1 \xi d \xi  \cdot \left(L_0 + L_1\left\|\omega^*\left(\theta_2\right)-\theta_2\right\|\right)\left\| \theta_1 -\theta_2\right\|^2\nonumber\\ 
     = &\frac{L_0 + L_1\left\|\omega^*\left(\theta_2\right)-\theta_2\right\|}{2}\left\| \theta_1 -\theta_2\right\|^2.
\end{align*}
This completes the proof.
\subsection{Proof of Lemma \ref{lemma: bound_hg}}\label{proof: lemma_bound_hg}
    By the definition of $h_g^t$ in (\ref{eq:g_grad}), we have
    \begin{align*}
        \left\|h_g^t\right\|=& 
        \Big\|\phi(s_t,a_t)\!\big(\phi(s_t,a_t)^{\!\top}\!\omega^t \!-\! R(s_t,a_t) 
        \!-\!\gamma \hat{G}_{\tau,\delta}(\hat{Q}_{\theta^t}\!\left(s^\prime_t,\cdot\right))\!\big)\!\Big\|\\
        \leq & \left\|\omega^t\right\|+ R_{\max} +\delta\\
         \leq &\frac{2\left(R_{\max} +\delta\right)}{\lambda_g},
    \end{align*}
where the last inequality is due to the projection operator in the update \eqref{eq: updating_omega} and the fact $\lambda_g \leq 1$.
\subsection{Proof of Lemma \ref{lemma: tracking_error}}\label{proof: lemma_tracking_error}
To start, we define 
\[
o^t_g(\theta,\omega) \triangleq \phi(s_t,a_t)(\phi(s_t,a_t)^\top\omega - R(s_t,a_t) -\gamma  \hat{G}_{\tau,\delta}(\hat{Q}_{\theta}(s^\prime_t,\cdot))),
\]
and 
\[
\xi^t_g(\theta,\omega) \triangleq \langle \omega^*(\theta)-\omega,o^t_g(\theta,\omega)  -\nabla_\omega g(\theta,\omega)\rangle.
\]
$\xi^t_g$ has the following properties.
\begin{lemma}\label{lemma: xi_g_property}
     For $t\geq0$, $\xi^t_g(\theta,\omega)$ is
   \begin{enumerate}
       \item $ \frac{8(R_{\max} + \delta)}{\lambda_g}$-Lipchitz in $\omega$ for $\omega\in\Omega$,
       \item and $\frac{4(R_{\max} + \delta)(1+\lambda_g)}{\lambda_g^2}$-Lipschitz in $\theta$.
   \end{enumerate}
\end{lemma}
The proof of Lemma \ref{lemma: xi_g_property} is in Appendix \ref{proof: lemma_xi_g_property}.

 For $t\geq T_\alpha$, following the update of $\omega^{t+1}$ in \eqref{eq: updating_omega} and taking the expectation yields  
\begin{align}\label{eq: omega_bound_1}
&\mathbb{E}\left[\left\|\omega^{*}(\theta^{t})-\omega^{t+1}\right\|^2\right]\nonumber\\
       \leq& \mathbb{E}\left[\left\|\omega^{*}\left(\theta^t\right)-\omega^{t}+\beta h_g^{t}\right\|^2\right]\nonumber\\
  = &\mathbb{E}\left[\left\|\omega^{*}\left(\theta^t\right)-\omega^{t}\right\|^2\right] + 2\beta\mathbb{E}\left[\left\langle\omega^{*}\left(\theta^t\right)-\omega^{t}, \nabla_\omega g(\theta^t,\omega^t) \right\rangle\right] \nonumber \\
  & + 2\beta \mathbb{E}\left[\xi^t_g(\theta^t,\omega^t)\right]+
    \beta^2 \mathbb{E}\left[\left\|h_g^{t}\right\|^2\right]\nonumber\\
    \leq & (1 - \beta\lambda_g)\mathbb{E}\left[\left\|\omega^{*}\left(\theta^t\right)-\omega^{t}\right\|^2\right] + 2\beta\mathbb{E}\left[\xi^t_g(\theta^t,\omega^t)\right] \nonumber\\
    &+ \beta^2 \mathbb{E}\left[\left\|h_g^{t}\right\|^2\right],
\end{align}
where the first inequality follows from the property of the projection $\mathcal{P}$ and the last inequality is due to Proposition~\ref{prop: g_convex}.

We first analyze the first term in (\ref{eq: omega_bound_1}). If $\beta \geq 1/\lambda_g$, this term can be upper bounded by $0$ and the subsequent results are easy to derive. If otherwise, meaning $\beta < 1/\lambda_g$, we have
\begin{align}\label{eq: omega_bound_term2}
   &(1-\beta\lambda_g) \mathbb{E}\left[\|\omega^{*}\left(\theta^t\right)-\omega^{t}\|^2\right] \nonumber \\
    &= \left(1-\beta\lambda_g\right)\mathbb{E}\left[\left\|\omega^{*}\left(\theta^t\right)-\omega^*(\theta^{t-1})+\omega^*(\theta^{t-1})-\omega^{t}\right\|^2\right]\nonumber\\
    &\leq  \left(1-\beta\lambda_g\right)\left(1+c\right)\mathbb{E}\left[\left\|\omega^{*}\left(\theta^t\right)-\omega^*(\theta^{t-1})\right\|^2\right]\nonumber\\
    & \ \ \ \ \ \ +\left(1-\beta\lambda_g\right)\left(1+\frac{1}{c}\right)\mathbb{E}\left[\left\|\omega^*(\theta^{t-1})-\omega^{t}\right\|^2\right]\nonumber\\
    &\leq  \left(1-\beta\lambda_g\right) \left( (1+c) \frac{4\alpha^2}{\lambda_g^2} + \left(1+\frac{1}{c}\right)\mathbb{E}\left[\left\|\omega^*(\theta^{t-1})-\omega^{t}\right\|^2\right]\right) \nonumber \\
    &\leq \frac{8 \alpha^2}{\beta \lambda_g^3} + \left(1 - \frac{\beta \lambda_g}{2}\right) \mathbb{E}\left[\left\|\omega^*(\theta^{t-1})-\omega^{t}\right\|^2\right],
\end{align}
where the first inequality is due to Young's inequality (and holds for any $c>0$), the second one follows from combining Lemma \ref{lemma: omega_star_lipschitz} and Lemma \ref{lemma: theta_update_bound}, and the last one is obtained by setting $c = \frac{2(1-\beta\lambda_g)}{\beta\lambda_g}$. 


For the second term in (\ref{eq: omega_bound_1}), we first define a random variable $\Tilde{\xi}_g(\theta,\omega)$:
\begin{align*}
    \Tilde{\xi}_g(\theta,\omega) \triangleq \langle \omega^*(\theta)-\omega, \phi(\tilde{s},\tilde{a})(\phi(\tilde{s},\tilde{a})^\top\omega - R(\tilde{s},\tilde{a}) -\gamma  \hat{G}_{\tau,\delta}(\hat{Q}_{\theta}(\tilde{s}^\prime,\cdot)))-\nabla_\omega g(\theta,\omega)\rangle,
\end{align*}
where $(\tilde{s},\tilde{a},\tilde{s}^\prime)\sim \mathcal{D}$.
Given $t\geq T_{\alpha}$, we have
\begin{align}\label{eq: omega_bound_term3}
    \mathbb{E}\left[\xi^t_g(\theta^t,\omega^t) \mid \mathcal{F}_{t-T_\alpha}\right]
    = & \mathbb{E}\left[\xi^t_g(\theta^t,\omega^t) -\xi^t_g(\theta^t,\omega^{t-T_\alpha})   \mid \mathcal{F}_{t-T_\alpha}\right] \nonumber\\
    &+ \mathbb{E}\left[\xi^t_g(\theta^t,\omega^{t-T_\alpha}) -\xi^t_g(\theta^{t-T_\alpha},\omega^{t-T_\alpha})   \mid \mathcal{F}_{t-T_\alpha}\right]\nonumber\\
    &+ \mathbb{E}\left[\xi^t_g(\theta^{t-T_\alpha},\omega^{t-T_\alpha}) -\tilde{\xi}_g(\theta^{t-T_\alpha},\omega^{t-T_\alpha})   \mid \mathcal{F}_{t-T_\alpha}\right]\nonumber\\
    \leq &  \frac{8(R_{\max} + \delta)}{\lambda_g} \mathbb{E}\left[\left\|\omega^t-\omega^{t-T_\alpha}\right\|\mid \mathcal{F}_{t-T_\alpha}\right]\nonumber\\
    & +  \frac{4(R_{\max} + \delta)(1+\lambda_g)}{\lambda_g^2}\mathbb{E}\left[\left\|\theta^t-\theta^{t-T_\alpha}\right\|\mid \mathcal{F}_{t-T_\alpha}\right]\nonumber\\
    & +  \frac{4\left(R_{\max} +  \delta\right)^2}{\lambda_g^2} d_{T V}\left(\mathbb{P}\left(\left(s_t, a_t,s_t^\prime\right) \in \cdot \mid s_{t-T_\alpha}, \pi_{\mathrm{bhv}},P\right), \mathcal{D} \right)\nonumber\\
    \leq & \frac{(16\beta T_\alpha + 4\alpha)(R_{\max} + \delta)^2 + 8\alpha T_\alpha(R_{\max} + \delta)(1+\lambda_g)}{\lambda_g^2},
\end{align}
where the first inequality is due to Lemma \ref{lemma: xi_g_property}, Assumption \ref{assump: Ergodicity}, and Lemma 1 in \cite{shen2020asynchronous}.

The last term in (\ref{eq: omega_bound_1}) can be bounded by Lemma~\ref{lemma: bound_hg} as 
\begin{equation}\label{eq: omega_bound_term4}
\mathbb{E}\left[\left\|h_g^{t}\right\|^2\right]\leq\frac{4\left(R_{\max} +  \delta\right)^2}{\lambda_g^2}.
\end{equation}

Taking the full expectation of \eqref{eq: omega_bound_term3} and substituting it with \eqref{eq: omega_bound_term2}, \eqref{eq: omega_bound_term4} into \eqref{eq: omega_bound_1} yields
\begin{align}\label{eq: omega_bound_3}
    \mathbb{E}\left[\left\|\omega^{*}(\theta^{t}) -\omega^{t+1}\right\|^2\right] 
    & \leq  \left(1-\frac{\beta\lambda_g}{2}\right)\mathbb{E}\left[\left\|\omega^*(\theta^{t-1})-\omega^{t}\right\|^2\right]+ \frac{8\alpha^2}{\beta\lambda_g^3}\nonumber\\
    &\quad +\frac{(32\beta^2 T_\alpha + 8\alpha\beta+4\beta^2)(R_{\max} + \delta)^2}{\lambda_g^2}\nonumber\\
    &+ \frac{16\alpha\beta T_\alpha(R_{\max} + \delta)(1+\lambda_g)}{\lambda_g^2}.
\end{align} 
Summing it from $t= T_\alpha+1$ to $t = T$ and rearranging the it gives
\begin{align}\label{eq: omega_bound_4}
    \frac1T \sum_{t=T_\alpha}^{T-1} \mathbb{E} \left\|\omega^{*}(\theta^{t})-\omega^{t+1}\right\|^2 
    \leq & \frac{2}{\beta\lambda_g T}\mathbb{E}\left\|\omega^{*}(\theta^{T_\alpha})-\omega^{T_\alpha+1}\right\|^2+ \frac{16\alpha^2}{\beta^2\lambda_g^4}\nonumber\\
    &+ \frac{(64\beta T_\alpha + 16\alpha+8\beta )(R_{\max} + \delta)^2}{\lambda_g^3}\nonumber\\ 
    &+ \frac{32\alpha T_\alpha (R_{\max} + \delta)(1+\lambda_g)}{\lambda_g^3}\nonumber\\
    = &\mathcal{O}\left(\frac{1}{\beta T} + \frac{\alpha^2}{\beta^2} + \beta T_\alpha + \alpha + \beta + \alpha T_\alpha\right).
\end{align}

Given the definition of $T_\alpha$ in (\ref{eq: T_alpha}) and the choice of $\alpha$, we obtain $T_\alpha =\mathcal{O}\left(\log T\right)$.
Thus, we can derive 
\begin{align}\label{eq: omega_bound_5}
    \sum_{t=0}^{T_\alpha-1}\mathbb{E}\left\|\omega^{*}(\theta^{t})-\omega^{t+1}\right\|^2 \leq T_\alpha \cdot \frac{4\left(R_{\max}+ \delta\right)^2}{\lambda_g^2}
    = \mathcal{O}\left(\log T \right).
\end{align}
Combining (\ref{eq: omega_bound_4}) and (\ref{eq: omega_bound_5}) yields
\begin{align*}
    \frac{1}{T}\sum_{t=0}^{T-1} \mathbb{E} \left\|\omega^{*}(\theta^{t})-\omega^{t+1}\right\|^2 
   \leq  & \mathcal{O}\left(\frac{\log T}{T} + \frac{1}{\beta T} + \frac{\alpha^2}{\beta^2} + \beta \log T + \alpha + \beta + \alpha \log T\right)\nonumber\\
    = &\mathcal{O}\left(\frac{\log T}{T^{1/2}}\right),
\end{align*}
where the last equality is given by $\alpha = \alpha T^{-3/4}$ and $\beta = \beta T^{-1/2}$.

\subsection{Proof of Lemma \ref{lemma: upper_error}}\label{proof: lemma_upper_error}
Define
\begin{align*}
    o^t_f(\theta,\omega) \triangleq\bigg(\gamma z\left(s^\prime_t; \tau,\delta,\theta\right)\sum_{a^\prime\in\mathcal{A}}\pi_{\theta}\left(a^\prime\mid s^\prime_t\right)\phi(s_t^\prime,a^\prime)
    -\phi(s_t,a_t)\bigg) \phi(s_t,a_t)^\top\left(\omega-\theta\right),
\end{align*}
and 
\[
\xi_f^t\left(\theta,\omega\right) \triangleq\left\langle \bar{\nabla}_\theta  f(\theta,\omega), \frac{1}{\|\omega-\theta\|}\left(\bar{\nabla}_\theta  f(\theta,\omega) -o^t_f(\theta,\omega)\right) \right\rangle.
\]
$\xi^t_g$ has the following properties.
\begin{lemma}\label{lemma: xi_f_property}
    For $t \geq 0$,
     \begin{enumerate}
    \item $\xi_f^t(\theta,\omega)$ is $24$-Lipschitz in $\omega$ for $\omega\in \Omega$;
    \item for  $\theta_1,\theta_2\in \mathbb{R}^d,\omega\in \Omega$,
    \begin{align*}
        \left|\xi^t_f(\theta_1,\omega) - \xi^t_f(\theta_2,\omega)\right|\leq \left( \frac{8L_1}{\sigma_{\min}}\|\bar{\nabla}_\theta f(\theta_1,\omega)\|+24\right)\left\|\theta_1-\theta_2\right\|.
    \end{align*}
\end{enumerate}
\end{lemma}
The proof for Lemma \ref{lemma: xi_f_property} is in Appendix \ref{proof: lemma_xi_f_property}.

We first consider the case of $\omega^{t+1}\ne \theta^t$. By Lemma \ref{lemma: J_smoothness}, for $t\geq T_{\alpha}$, we have
\begin{align}\label{eq: theta_bound}
    J\left(\theta^{t+1}\right) -J\left(\theta^{t}\right)
    \leq &\left\langle \nabla J\left(\theta^{t}\right), \theta^{t+1} - \theta^t\right\rangle + \frac{L_0 + L_1\left\|\omega^*\left(\theta^t\right)-\theta^t\right\|}{2}\left\| \theta^{t+1} -\theta^t\right\|^2\nonumber\\
    \leq & \left\langle \nabla J\left(\theta^{t}\right), -\frac{\alpha  }{\|\omega^{t+1}-\theta^t\|} {h}_f^t\right\rangle +2\alpha ^2L_0 + 2\alpha ^2L_1\left\|\omega^*\left(\theta^t\right)-\theta^t\right\|\nonumber\\
    \leq &- \alpha \frac{\left\|\bar{\nabla}_\theta  f(\theta^t,\omega^{t+1})\right\|^2}{\left\|\omega^{t+1}-\theta^t\right\|} + \left\langle \nabla J\left(\theta^{t}\right)-\bar{\nabla}_\theta f(\theta^t,\omega^{t+1}), -\frac{\alpha }{\|\omega^{t+1}-\theta^t\|} h_f^t\right\rangle\nonumber\\
    &+2\alpha ^2L_0 + 2\alpha ^2L_1\left\|\omega^*\left(\theta^t\right)-\theta^t\right\|,
\end{align}
where the second inequality follows from Lemma \ref{lemma: theta_update_bound}. Under Assumption \ref{assump: non-singular}, the $\ell_2$-norm of the surrogate gradient is lower bounded as
\begin{equation}\label{eq: grad_f_lb}
\left\|\bar{\nabla}_\theta f\left(\theta^t,\omega^{t+1}\right) \right\| \geq 
\left\|\omega^{t+1}-\theta^t\right\|.
\end{equation}
Thus we can bound the first term on the right-hand side of \eqref{eq: theta_bound} as
\begin{align}\label{eq: theta_bound_term1}
    - \alpha \frac{\left\|\bar{\nabla}_\theta  f(\theta^t,\omega^{t+1})\right\|^2}{\left\|\omega^{t+1}-\theta^t\right\|} \leq -\alpha \sigma_{\min} \left\|\bar{\nabla}_\theta  f(\theta^t,\omega^{t+1})\right\|.
\end{align}

For the second term on the right-hand side of \eqref{eq: theta_bound},
\begin{align}\label{eq: theta_bound_term2}
   &\left\langle \nabla J\left(\theta^{t}\right)-\bar{\nabla}_\theta f(\theta^t,\omega^{t+1}), -\frac{\alpha }{\|\omega^{t+1}-\theta^t\|} h_f^t\right\rangle \nonumber\\ 
    \leq &\alpha\left\|\nabla J\left(\theta^{t}\right) - \bar{\nabla}_\theta f\left(\theta^t,\omega^{t+1}\right)\right\|\frac{\left\| h_f^t \right\|}{\|\omega^{t+1}-\theta^t\|}\nonumber\\
    \leq &4\alpha\left\|\omega^*\left(\theta^t\right) - \omega^{t+1}\right\|,
\end{align}
where the last inequality is due to Lemma \ref{lemma: gradient_bound}.

Next, we focus on the third term on the right-hand side of \eqref{eq: theta_bound}. We define 
\begin{align*}
    \tilde{\xi}_f(\theta,\omega) =  \bigg\langle   \frac{\bar{\nabla}_\theta f(\theta,\omega)}{\|\omega-\theta\|}, \Big(\bar{\nabla}_\theta  f(\theta,\omega)- \gamma z\left(\tilde{s}^\prime; \tau,\delta,\theta\right)\big(\sum_{a^\prime\in\mathcal{A}}\pi_{\theta}\left(a^\prime\mid \tilde{s}^\prime\right)\phi(\tilde{s}^\prime,a^\prime)
    -\phi(\tilde{s},\tilde{a})\big) \phi(\tilde{s},\tilde{a})^\top\left(\omega-\theta\right)\Big)\bigg\rangle,
\end{align*}
where $(\tilde{s},\tilde{a},\tilde{s}^\prime)\sim \mathcal{D}$.
For $t\geq T_{\beta}+1$, by taking the conditional expectation on $\mathcal{F}_t^\prime$, we obtain
\begin{align}\label{eq: theta_bound_term3}
      & \mathbb{E}\left[\xi^t_f(\theta^t,\omega^{t+1})\mid \mathcal{F}^\prime_{t-T_\beta-1}\right]\nonumber\\
   \leq &  \mathbb{E}\left[\xi^t_f(\theta^t,\omega^{t+1}) - \xi^t_f(\theta^{t-T_\beta-1},\omega^{t+1})\mid \mathcal{F}^\prime_{t-T_\beta-1}\right]\nonumber\\
   &+  \mathbb{E}\left[\xi^t_f(\theta^{t-T_\beta-1},\omega^{t+1}) - \xi^t_f(\theta^{t-T_\beta-1},\omega^{t-T_\beta})\mid \mathcal{F}^\prime_{t-T_\beta-1}\right]\nonumber\\
   & +  \mathbb{E}\left[\xi^t_f(\theta^{t-T_\beta-1},\omega^{t-T_\beta}) - \tilde{\xi}_f(\theta^{t-T_\beta-1},\omega^{t-T_\beta})\mid \mathcal{F}^\prime_{t-T_\beta-1}\right]\nonumber\\
\leq & \alpha (T_\beta+1) \left(\frac{16L_1}{\sigma_{\min}}\mathbb{E}\left[\|\bar{\nabla}_\theta f(\theta^t,\omega^{t+1})\|\mid \mathcal{F}^\prime_{t-T_\beta-1}\right]+48\right)\nonumber\\
& + 48\beta (T_\beta+1) (R_{\max}+\delta)/\lambda_g\nonumber\\
& + 2\left\|\bar{\nabla}_\theta  f(\theta^{t-T_\beta-1},\omega^{t-T_\beta})\right\|d_{T V}\left(\mathbb{P}\left(\left(s_t, a_t,s_t^\prime\right) \in \cdot \mid s_{t-T_\beta}, \pi_{\mathrm{bhv}},P\right), \mathcal{D} \right)\nonumber\\
\leq & \alpha (T_\beta+1) \left(\frac{16L_1}{\sigma_{\min}}\mathbb{E}\left[\|\bar{\nabla}_\theta f(\theta^t,\omega^{t+1})\|\mid \mathcal{F}^\prime_{t-T_\beta-1}\right]+48\right)\nonumber\\
& + 48\beta (T_\beta+1) (R_{\max}+\delta)/\lambda_g\nonumber\\
& + 2\beta \mathbb{E}\left[\left\|\bar{\nabla}_\theta  f(\theta^{t-T_\beta-1},\omega^{t-T_\beta}) -\bar{\nabla}_\theta  f(\theta^{t-T_\beta-1},\omega^{t+1})\right\|\right]\nonumber\\
&+2\beta\mathbb{E}\left[\left\|\bar{\nabla}_\theta  f(\theta^{t-T_\beta-1},\omega^{t+1})-\bar{\nabla}_\theta  f(\theta^{t},\omega^{t+1})\right\|\mid \mathcal{F}^\prime_{t-T_\beta-1} \right]\nonumber\\
& +2\beta \mathbb{E}\left[\left\|\bar{\nabla}_\theta  f(\theta^{t},\omega^{t+1})\right\|\mid \mathcal{F}^\prime_{t-T_\beta-1} \right]\nonumber\\
\leq & \left((16+4\beta)\alpha (T_\beta+1) \frac{L_1}{\sigma_{\min}}+ 2\beta\right) \mathbb{E}\left[\left\|\bar{\nabla}_\theta  f(\theta^{t},\omega^{t+1})\right\|\mid \mathcal{F}^\prime_{t-T_\beta-1} \right]\nonumber\\
& + (48+8\beta)\alpha (T_\beta+1) + 48\beta (T_\beta+1) (R_{\max}+\delta)/\lambda_g.
\end{align}

The last term in \eqref{eq: theta_bound} can be bounded as 
\begin{align*}
    &2\alpha^2L_1\left\|\omega^*(\theta^t)-\theta^t\right\|\nonumber\\
    \leq & 2\alpha^2L_1\left\|\omega^*(\theta^t) - \omega^{t+1}\right\| +2\alpha^2L_1\left\|\theta^t - \omega^{t+1}\right\| \nonumber \\
    \leq & 2\alpha^2L_1\left\|\omega^*(\theta^t) - \omega^{t+1}\right\| +2\alpha^2L_1\frac{\left\|\bar{\nabla}_\theta f(\theta^t,\omega^{t+1})\right\|}{\sigma_{\min}},
\end{align*}
where the last inequality follows from~\eqref{eq: grad_f_lb}.

Taking full expectation in \eqref{eq: theta_bound} and plugging \eqref{eq: theta_bound_term1}, \eqref{eq: theta_bound_term2}, \eqref{eq: theta_bound_term3} and the preceding inequality yield 
\begin{align*}
    &\mathbb{E}\left[J(\theta^{t+1})-J(\theta^t)\right]\\ \leq & -\alpha ( \sigma_{\min}- ((16\alpha (T_\beta+1) +  4\alpha\beta (T_\beta+1) + 2\alpha^2)\frac{L_1}{\sigma_{\min}}  + 2\beta))\nonumber\\
    &\times \mathbb{E}\left[\left\|\bar{\nabla}_\theta f(\theta^t,\omega^{t+1})\right\|\right]\\
    & + (4\alpha +2\alpha^2 L_1)\mathbb{E}\left\|\omega^*(\theta^t) - \omega^{t+1}\right\|+ 48\alpha^2 (T_\beta+1)  \\
    &+ 8\alpha^2\beta (T_\beta+1)+ 48\alpha\beta (T_\beta+1) (R_{\max}+\delta)/\lambda_g + 2\alpha^2 L_0.
\end{align*}
Observe that the above inequality also holds for the case of $\omega^{t+1} = \theta^t$. 

Given the conditions 
\begin{align*}
    \alpha \leq \frac{\beta}{\frac{\log(\beta/\kappa)}{\log \rho}+2},
    \beta \leq&  \frac{\sigma_{\min}}{4},\\
    64\beta+16\beta^2 +8\alpha^2 \leq&   \frac{\sigma_{\min}^2}{L_1},
\end{align*}
we have 
 \begin{align*}
    &\mathbb{E}\left[J(\theta^{t+1})-J(\theta^t)\right]\nonumber\\
    \leq & -\frac{\alpha \sigma_{\min}}{4} \mathbb{E}\left[\left\|\bar{\nabla}_\theta f(\theta^t,\omega^{t+1})\right\|\right] + (4\alpha +2\alpha^2 L_1) \mathbb{E}\left\|\omega^*(\theta^t) - \omega^{t+1}\right\|\\
    &+ 48\alpha^2 (T_\beta+1) + 8\alpha^2\beta (T_\beta+1)\nonumber\\
    &+ 48\alpha\beta (T_\beta+1) (R_{\max}+\delta)/\lambda_g + 2\alpha^2 L_0.
\end{align*}

Summing this inequality from $t = T_\beta+1$ to $t =T-1$ gives:
\begin{align}\label{eq: theta_bound_2}
    &\frac{\alpha \sigma_{\min}}{4} \sum_{t=T_\beta+1}^{T-1}\mathbb{E}\left[\left\|\bar{\nabla}_\theta f(\theta^t,\omega^{t+1})\right\|\right]\nonumber\\ \leq & \mathbb{E}\left[ J(\theta^{T_\beta+1})\right] + (4\alpha +2\alpha^2 L_1)\sum_{t=T_\beta+1}^{T-1} \mathbb{E}\left\|\omega^*(\theta^t) - \omega^{t+1}\right\|\nonumber\\
    & + 48\alpha^2 (T_\beta+1) T + 8\alpha^2\beta (T_\beta+1)  T\nonumber\\
    &+ 48\alpha\beta (T_\beta+1) T (R_{\max}+\delta)/\lambda_g + 2\alpha^2 TL_0.
\end{align}

To continue, we consider the first term in \eqref{eq: theta_bound_2}. By the triangle inequality and Lemma \ref{lemma: theta_update_bound}, we have
\begin{align*}
    \left\| \theta^{T_{\beta}+1}\right\| \leq \sum_{t = 0}^{T_{\beta}}\left\|\theta^{t+1} - \theta^t\right\| + \left\|\theta^0\right\|
    \leq  2\alpha (T_{\beta}+1) + \left\|\theta^0\right\|.
\end{align*}

Given the definition of the objective function $J$ in \eqref{eq: bi}, we obtain
\begin{align}\label{eq: theta_bound2_term1}
&\mathbb{E} J\left(\theta^{T_{\beta}+1}\right)\nonumber\\
\leq& \left\|\omega^*\left(\theta^{T_{\beta}+1}\right) - \theta^{T_{\beta}+1}\right\|^2\nonumber\\
\leq & 2 \left\|\omega^*\left(\theta^{T_{\beta}+1}\right)\right\|^2 + 2\left\|\theta^{T_{\beta}+1}\right\|^2\nonumber\\
\leq & \frac{2\left(R_{\max}+\delta\right)^2}{\lambda_g^2} + 16\alpha^2 (T_{\beta}+1)^2  + 4\left\|\theta^0\right\|^2,
\end{align}
where the last two inequalities follow from Young's inequality.

For the second term in \eqref{eq: theta_bound_2}, we bound it by using \eqref{eq: omega_bound_final}:
\begin{align}\label{eq: theta_bound2_term2}
     \sum_{t=T_{\beta}+1}^{T-1}\mathbb{E}\left\|\omega^*\left(\theta^t\right) - \omega^{t+1}\right\| 
    = \mathcal{O}( T^{3/4}(\log T)^{1/2}).
\end{align}

Substituting \eqref{eq: theta_bound2_term1} and \eqref{eq: theta_bound2_term2} into \eqref{eq: theta_bound_2} and dividing $\alpha\sigma_{\min}/4$ on both sides yield that
\begin{align}\label{eq: theta_bound_3}
    &\sum_{t= T_{\beta}+1}^{T-1}\mathbb{E} \left\|\bar{\nabla}_\theta f\left(\theta^t,\omega^{t+1}\right)\right\|\nonumber\\
    \leq &  \frac{8\left(R_{\max}+\delta\right)^2}{\alpha\sigma_{\min}\lambda_g^2} + \frac{64\alpha (T_\beta+1)^2 }{\sigma_{\min}} + \frac{16\left\|\theta^0\right\|^2}{\alpha\sigma_{\min}}\nonumber\\
    &+ \frac{192\alpha (T_\beta+1) T}{\sigma_{\min}}  + \frac{32\alpha \beta (T_\beta+1)  T }{\sigma_{\min}}\nonumber\\
    &+ \frac{192\beta (T_\beta+1) T (R_{\max}+\delta)}{\lambda_g \sigma_{\min}} \nonumber\\
    &+\frac{8\alpha T  L_0}{\sigma_{\min}} +\mathcal{O}( T^{3/4}(\log T)^{1/2}).
\end{align}

For $t\leq T_\beta$, we can bound the $\ell_2$-norm of the surrogate gradient as
\begin{align*}
    \left\|\bar{\nabla}f\left(\theta^t,\omega^{t+1}\right) \right\| \leq &  \left\|\hat{\Sigma}_{\tau,\delta,\theta^t}\right\| \left\|\omega^{t+1}-\theta^t\right\|\nonumber\\
    \leq & 2\left(\left\|\omega^{t+1}\right\|+\left\|\theta^t\right\|\right)\nonumber\\
    \leq &2\left(\frac{R_{\max}+\delta}{\lambda_g} + 2\alpha T_{\beta}+\left\|\theta^0\right\| \right),
\end{align*}
where the last inequality follows from Lemma \ref{lemma: theta_update_bound}. As a result, we have
\begin{align}\label{eq: theta_bound_4}
    \sum_{t= 0}^{T_{\beta}-1}\left\|\bar{\nabla}f\left(\theta^t,\omega^{t+1}\right) \right\| \leq 2\left(\frac{R_{\max}+\delta}{\lambda_g} +2\alpha T_{\beta}+\left\|\theta^0\right\| \right) T_{\beta}
    = \mathcal{O}\left(\log T \right).
\end{align}
Combining \eqref{eq: theta_bound_3} and \eqref{eq: theta_bound_4}, and dividing $T$ lead to
\begin{align*}
       \frac{1}{T}\sum_{t= 0}^{T-1} \left\|\bar{\nabla}_\theta f\left(\theta^t,\omega^{t+1}\right) \right\|
    &\leq \mathcal{O}\bigg(\frac{\log T}{T} + \frac{1}{\alpha T} + \frac{\alpha (\log T)^2}{T} + \alpha \log T + \alpha\beta \log T\\
    &\quad+ \beta \log T + \alpha + \frac{(\log T)^{1/2}}{T^{1/4}}\bigg)\\
    & = \mathcal{O}(\frac{(\log T)^{1/2}}{T^{1/4}}),
\end{align*}
where the last equality is given by $\alpha = \alpha T^{-3/4}$ and $\beta = \beta T^{-1/2}$.

\subsection{Proof of Lemma \ref{lemma: xi_g_property}}\label{proof: lemma_xi_g_property}
We start with the first property.
    For $\omega_1,\omega_2\in\Omega,\theta\in \mathbb{R}^d$, it has 
    \begin{align*}
       &\left |\xi^t_g(\theta,\omega_1)-\xi^t_g(\theta,\omega_2)\right|\\ \leq &\left |\left\langle \omega^*(\theta)-\omega_2, \left(\phi(s_t,a_t)\phi(s_t,a_t)^\top-\mathbb{E}_{ \mathcal{D}}\left[\phi(s,a)\phi(s,a)^\top\right]\right) \left(\omega_1-\omega_2\right) \right\rangle\right|\\
        & + \left|\left\langle\omega_1-\omega_2,o^t_g(\theta,\omega_1)-\nabla_\omega g(\theta,\omega_1)\right\rangle\right|\\
        \leq & \frac{8(R_{\max} + \delta)}{\lambda_g} \left\|\omega_1-\omega_2\right\|.
    \end{align*}
    Next, we prove the second property. For $\omega\in\Omega, \theta_1,\theta_2\in\mathbb{R}^d$, it follows that
    \begin{align*}
         &\left |\xi^t_g(\theta_1,\omega)-\xi^t_g(\theta_2,\omega)\right|\\ \leq & \left|\left\langle \omega-\omega^*(\theta_1), \phi(s_t,a_t)\left(\hat{G}_{\tau,\delta}(\hat{Q}_{\theta_1}(s^\prime_t,\cdot))-\hat{G}_{\tau,\delta}(\hat{Q}_{\theta_2}(s^\prime_t,\cdot))\right) \right\rangle\right|\\
         &+\left|\left\langle \omega-\omega^*(\theta_1), \mathbb{E}_{\mathcal{D}}\left[\phi(s,a)\left(\hat{G}_{\tau,\delta}(\hat{Q}_{\theta_1}(s^\prime,\cdot))-\hat{G}_{\tau,\delta}(\hat{Q}_{\theta_2}(s^\prime,\cdot))\right) \right]\right\rangle\right|\\
         & + \left|\left\langle\omega^*(\theta_1)-\omega^*(\theta_2), o^t_g(\theta_2,\omega)-\nabla_\omega g(\theta_2,\omega)  \right\rangle\right|\\
         \leq & \frac{4(R_{\max} + \delta)(1+\lambda_g)}{\lambda_g^2} \left\|\theta_1-\theta_2\right\|,
    \end{align*}
    where the last inequality follows from 
    \[
    \left\|\nabla_\theta \hat{G}_{\tau,\delta}(\hat{Q}_\theta(s,\cdot))\right\| = \left\|z(s;\tau,\delta,\theta)\sum_{a}\pi_\theta(a\mid s)\phi(s,a)\right\| \leq 1,
    \]
    and Lemma \ref{lemma: omega_star_lipschitz}.
\subsection{Proof of Lemma \ref{lemma: xi_f_property}}\label{proof: lemma_xi_f_property}
 We start with the first property. For $\omega_1,\omega_2\in \Omega,\theta\in\mathbb{R}^d$, it has
    \begin{align*}
       & \left|\xi_f^t(\theta,\omega_1)-\xi_f^t(\theta,\omega_2)\right| \\\leq& \left|\left\langle \bar{\nabla}_\theta  f(\theta,\omega_1)-\bar{\nabla}_\theta f(\theta,\omega_2), \frac{\bar{\nabla}_\theta  f(\theta,\omega_1)-o^t_f(\theta,\omega_1)}{\|\omega_1-\theta\|}\right\rangle\right| \\
    & + \left|\left\langle\bar{\nabla}_\theta f(\theta,\omega_2),\frac{\bar{\nabla}_\theta  f(\theta,\omega_1)}{\|\omega_1-\theta\|} -\frac{\bar{\nabla}_\theta  f(\theta,\omega_2)}{\|\omega_2-\theta\|}  \right\rangle\right|\\
     & + \left|\left\langle\bar{\nabla}_\theta f(\theta,\omega_2),\frac{o^t_f(\theta,\omega_1)}{\|\omega_1-\theta\|} -\frac{o^t_f(\theta,\omega_2)}{\|\omega_2-\theta\|}  \right\rangle\right|\\
    \leq & 8\left\|\omega_1-\omega_2\right\| \\
    &+ \left|\left\langle\bar{\nabla}_\theta f(\theta,\omega_2), \frac{\|\omega_2-\theta\|\bar{\nabla}_\theta  f(\theta,\omega_1) -\|\omega_1-\theta\|\bar{\nabla}_\theta  f(\theta,\omega_2) }{\|\omega_1-\theta\|\|\omega_2-\theta\|}\right\rangle\right|\\
    & + \left|\left\langle \bar{\nabla}_\theta  f(\theta,\omega_2), \frac{\|\omega_2-\theta\|o^t_f(\theta,\omega_1) -\|\omega_1-\theta\|o^t_f(\theta,\omega_2) }{\|\omega_1-\theta\|\|\omega_2-\theta\|}\right\rangle\right|.
    \end{align*}
    The second term on the right-hand side of the above inequality can be bounded as follows, with a similar derivation applying to the third term.
    \begin{align*}
       & \left|\left\langle\bar{\nabla}_\theta f(\theta,\omega_2), \frac{\|\omega_2-\theta\|\bar{\nabla}_\theta  f(\theta,\omega_1) -\|\omega_1-\theta\|\bar{\nabla}_\theta  f(\theta,\omega_2) }{\|\omega_1-\theta\|\|\omega_2-\theta\|}\right\rangle\right|\\
       \leq & \left|\left\langle\frac{\bar{\nabla}_\theta f(\theta,\omega_2)}{\|\omega_2-\theta\|}, \frac{\|\omega_2-\theta\|-\|\omega_1-\theta\|}{\|\omega_1-\theta\|}\bar{\nabla}_\theta  f(\theta,\omega_1)\right\rangle\right|\\& + \left|\left\langle\frac{\bar{\nabla}_\theta f(\theta,\omega_2)}{\|\omega_2-\theta\|},\bar{\nabla}_\theta  f(\theta,\omega_1)-\bar{\nabla}_\theta  f(\theta,\omega_2)\right\rangle\right|\\
       \leq & 8 \left\|\omega_1-\omega_2\right\|
    \end{align*}
    Consequently, the proof is completed as 
    \begin{equation*}
         \left|\xi_f^t(\theta,\omega_1)-\xi_f^t(\theta,\omega_2)\right| \leq 24 \left\|\omega_1-\omega_2\right\|.
    \end{equation*}
    
 Regarding the second property of $\xi^t_f$, we derive the following inequality first. For $\theta_1,\theta_2\in\mathbb{R}^d,\omega\in\Omega$,
    \begin{align*}
        & \left\|\bar{\nabla}_\theta  f(\theta_1,\omega) -\bar{\nabla}_\theta  f(\theta_2,\omega) \right\|\\
        \leq &  \left\|\hat{\Sigma}_{\tau,\delta,\theta_1} (\omega-\theta_1)-\hat{\Sigma}_{\tau,\delta,\theta_2} (\omega-\theta_2) \right\|\\
        \leq & \left\|\left(\hat{\Sigma}_{\tau,\delta,\theta_1}-\hat{\Sigma}_{\tau,\delta,\theta_2}\right)(\omega-\theta_1)\right\|\\
        & + \left\|\hat{\Sigma}_{\tau,\delta,\theta_2}\left(\theta_1-\theta_2\right)\right\|\\
        \leq &L_1 \left\| \omega-\theta_1\right\| \left\|\theta_1-\theta_2\right\| + 2 \left\|\theta_1-\theta_2\right\|\\
        \leq & \frac{1}{\sigma_{\min}}L_1  \left\|\bar{\nabla}_\theta f(\theta_1,\omega)\right\| \left\|\theta_1-\theta_2\right\| + 2 \left\|\theta_1-\theta_2\right\|,
    \end{align*}
    where the third inequality follows from Lemma \ref{lemma: Sigma_hat_lipcshitz} and the last inequality is due to Assumption \ref{assump: non-singular}.  Consequently, for $\theta_1,\theta_2\in \mathbb{R}^d,\omega\in \Omega$, it has
    \begin{align*}
       & \left\|\xi^t_f(\theta_1,\omega)-\xi^t_f(\theta_2,\omega) \right\|\\
       \leq & \left|\left\langle\bar{\nabla}_\theta  f(\theta_1,\omega) -\bar{\nabla}_\theta  f(\theta_2,\omega) ,\frac{\bar{\nabla}_\theta f(\theta_1,\omega)-o^t_f(\theta_1,\omega)}{\|\omega-\theta_1\|}\right\rangle\right|\\
       & + \left|\left\langle \bar{\nabla}_\theta f(\theta_2,\omega), \frac{\bar{\nabla}_\theta  f(\theta_1,\omega)}{\|\omega-\theta_1\|} -\frac{\bar{\nabla}_\theta  f(\theta_2,\omega)}{\|\omega-\theta_2\|}\right\rangle\right|\\
        & + \left|\left\langle \bar{\nabla}_\theta f(\theta_2,\omega), \frac{o^t_f(\theta_1,\omega)}{\|\omega-\theta_1\|} -\frac{\bar{\nabla}_\theta  o^t_f(\theta_2,\omega)}{\|\omega-\theta_2\|}\right\rangle\right|\\
        \leq & \left(\frac{8L_1}{\sigma_{\min}}\|\bar{\nabla}_\theta f(\theta_1,\omega)\|+24\right)\left\|\theta_1-\theta_2\right\|.
    \end{align*}
\section{Proofs of Corollaries}
\subsection{Proof of Corollary \ref{corollary: target_convergence}}\label{proof: corollary_target_convergence}
The proof of the result in Corollary \ref{corollary: target_convergence} is straightforward. Given the triangle inequality and Lemma \ref{lemma: bound_hg}, we have
\begin{align*}
    \left\|\theta^t-\omega^t\right\| \leq& \left\|\theta^t-\omega^{t+1}\right\| + \left\|\omega^t-\omega^{t+1}\right\|\\
     = & \left\|\theta^t-\omega^{t+1}\right\| + \beta\left\|h_g^t\right\|\\
     \leq & \left\|\theta^t-\omega^{t+1}\right\| + \frac{2\beta \left(R_{\max} +  \delta\right)}{\lambda_g}.
\end{align*}
As a result, under Assumption \ref{assump: non-singular}, we can derive
\begin{align*}
    \frac{1}{T} \sum_{t= 0}^{T-1} \mathbb{E} \left\|\theta^t-\omega^t\right\| 
    &\leq \frac{1}{T}\sum_{t= 0}^{T-1}\mathbb{E} \left\|\theta^t-\omega^{t+1}\right\| + \frac{2\beta \left(R_{\max} +  \delta\right)}{\lambda_g}\\
    &\leq  \frac{1}{\sigma_{\min} T}\sum_{t= 0}^{T-1}\mathbb{E}\left\|\bar{\nabla}_\theta f\left(\theta^t,\omega^{t+1}\right)\right\| + \frac{2\beta \left(R_{\max} +  \delta\right)}{\lambda_g}\\
    &= \mathcal{O}\left(T^{-1/4}\right),
\end{align*}
where the last equality is from the Theorem~\ref{theorem: convergence}.
\subsection{Proof of Corollary \ref{corollary: policy_convergence}}\label{proof: corollary_policy_convergence}
\begin{proof}
    For \( t \geq 0 \), we have
    \begin{align*}
        \left\|\pi^*_{\tau} - \pi_{\theta^t} \right\|_{\infty} &= \max_{s\in\mathcal{S}} \|\pi^*_{\tau}(\cdot \mid s) - \pi_{\theta^t}(\cdot \mid s)\|_\infty\\
        &\leq \max_{s\in\mathcal{S}} \|\pi^*_{\tau}(\cdot \mid s) - \pi_{\theta^t}(\cdot \mid s)\|_2\\
        &\leq \frac{L_G}{\tau} \max_{s\in\mathcal{S}} \|Q^*_{\tau}(s,\cdot) - \hat{Q}_{\theta^t}(s,\cdot)\|_2\\
        &\leq \frac{\sqrt{|\mathcal{A}|}L_G}{\tau} \max_{s\in\mathcal{S}} \|Q^*_{\tau}(s,\cdot) - \hat{Q}_{\theta^t}(s,\cdot)\|_\infty\\
        &= \frac{\sqrt{|\mathcal{A}|}L_G}{\tau} \left\|Q^*_{\tau} - \hat{Q}_{\theta^t} \right\|_\infty.
    \end{align*}
Summing this inequality over \( t = 0 \) to \( t = T-1 \) and taking the expectation, we obtain
\[
    \sum_{t=0}^{T-1} \mathbb{E} \left\|\pi^*_{\tau} - \pi_{\theta^t} \right\|_{\infty} \leq \frac{\sqrt{|\mathcal{A}|}L_G}{\tau} \sum_{t=0}^{T-1} \mathbb{E} \left\|Q^*_{\tau} - \hat{Q}_{\theta^t} \right\|_\infty.
\]
The proof follows by applying Theorem \ref{theorem: estimation_bound}.
\end{proof}
\section{Proofs of Propositions}
\subsection{Proof of Proposition \label{prop: g_convex}}\label{proof: prop_g_convex}
Under Assumption \ref{assump: Ergodicity}, and with the linear independency of the basis vectors, the matrix $\Sigma$ is positive-definite. Consequently, there exists a positive constant $\lambda_g$ that lower-bounds the eigenvalues of $\Sigma$. 
For $\omega_1, \omega_2\in \Omega$, by Taylor theorem, we have
\begin{align}
    g(\theta,\omega_1) =& g(\theta,\omega_2) + \left\langle\nabla_{\omega}g(\theta,\omega_2),\omega_1-\omega_2 \right\rangle + \frac{1}{2}\left(\omega_1-\omega_2\right)^\top \Sigma \left(\omega_1-\omega_2\right) \nonumber\\
    \geq& g(\theta,\omega_2) + \left\langle\nabla_{\omega}g(\theta,\omega_2),\omega_1-\omega_2 \right\rangle + \frac{\lambda_g}{2}\left\|\omega_1-\omega_2\right\|^2.
\end{align}
The proof is complete.
\end{appendices}
\end{document}